%% file: main.tex
\documentclass[11pt]{article}
\include{newcommands}
\title{\vspace{1cm}\bf Gaussian Process Autonomous Mapping and Exploration for Range Sensing Mobile Robots\thanks{Accepted for Autonomous Robots. 
    \url{mani.ghaffari@gmail.com} -- \url{http://maanighaffari.com}}}
\author{Maani Ghaffari Jadidi\thanks{College of Engineering, University of Michigan, Ann Arbor, MI 48109 USA} \and
Jaime Valls Miro\thanks{Centre for Autonomous Systems (CAS), University
of Technology Sydney, NSW 2007, Australia} \and  Gamini Dissanayake\footnotemark[3]}

\begin{document}

\maketitle
\thispagestyle{empty}
\vspace{1.5cm}
\begin{abstract}
Most of the existing robotic exploration schemes use occupancy grid representations and geometric targets known as frontiers. The occupancy grid representation relies on the assumption of independence between grid cells and ignores structural correlations present in the environment. We develop a Gaussian Processes (GPs) occupancy mapping technique that is computationally tractable for online map building due to its incremental formulation and provides a continuous model of uncertainty over the map spatial coordinates. The standard way to represent geometric frontiers extracted from occupancy maps is to assign binary values to each grid cell. We extend this notion to novel probabilistic frontier maps computed efficiently using the gradient of the GP occupancy map. We also propose a mutual information-based greedy exploration technique built on that representation that takes into account all possible future observations. A major advantage of high-dimensional map inference is the fact that such techniques require fewer observations, leading to a faster map entropy reduction during exploration for map building scenarios. Evaluations using the publicly available datasets show the effectiveness of the proposed framework for robotic mapping and exploration tasks.
\end{abstract}
\clearpage
\thispagestyle{empty}
\tableofcontents
\clearpage

\section{Introduction}
\label{intro}
Exploring an unknown environment without any prior knowledge gives rise to difficulties for the robot to make sequential decisions that maximize the long-term expected reward or \emph{information gain}. Among these difficulties, available information in the current state of the robot is limited to its perception field and the partially known state of its trajectory and the map as \emph{a priori}. This leads the problem towards the sequential decision making under imperfect state information which is known to be NP-hard~\citep{singh2009efficient}.

Autonomous mobile robots are required to generate a spatial representation of the robot environment, this is known as the mapping problem. Solving this problem is an integral part of all autonomous navigation systems as the map encapsulates the knowledge of the robot about its surrounding. In robotic navigation tasks, a representation (map) that indicates occupied areas of the environment is required. Furthermore, it is desirable that such maps be generated autonomously where the robot explores new regions of an unknown environment. This is known as the autonomous exploration problem in robotics. In this article, we are concerned with autonomous exploration for map building when the robot pose is estimated by an appropriate strategy such as Pose SLAM~\citep{ila2010information}.

We develop a Gaussian Processes (GPs) occupancy mapping algorithm that is tailored for robotic navigation and is computationally tractable due to its incremental formulation. This representation has been shown to be superior to the traditional occupancy grid map~\citep{moravec1985high,elfes1987sonar} as it captures structural correlations present in the environment and produces a continuous representation of the sensing uncertainty in the map space~\citep{o2009contextual,t2012gaussian,kim2012building,kim2013continuous,kim2013occupancy,jadidi2013exploration,jadidi2013acra,maani2014com,maani2015mi,kim2015gpmap,jadidi2017warped}. Furthermore, this representation has applications beyond the occupancy mapping problem and it has been applied for large-scale terrain modeling~\citep{vasudevan2009gaussian}, active learning~\citep{krause2007nonmyopic}, building maps to predict the fluid concentration~\citep{stachniss2008gas}, informative path planning~\citep{binney2012branch}, robotic information gathering~\citep{hollinger2014sampling,hollinger2015long,jadidi2016sampling}, and high-dimensional semantic map representation~\citep{gpsm2017rssws}.

In the problem of robotic exploration for map building, targets are usually defined by geometric \emph{frontiers} extracted from the Occupancy Grid Map (OGM)~\citep{yamauchi1997frontier}. We propose an algorithm to extract frontiers from Gaussian Processes Occupancy Maps (GPOMs) representations in the form of a probability map. Furthermore, we develop an algorithm to numerically calculate the Mutual Information (MI) between the map and future measurements on that representation. MI is a measure of the value of information that quantifies the information gain from sensor measurements~\citep{Krause:uai05infogain}. The maximum expected utility principle states that the robot should choose the action that maximizes its expected utility, in the current state~\citep[page 483]{russell2009artificial}. The expectation is taken due to the stochastic nature of the state and observations. The proposed MI algorithm takes into account all possible future measurements (by taking expectations over them), and therefore, is a suitable utility function.

The MI-based utility function is computed at the centroids of geometric frontiers and the frontier with the highest information gain is chosen as the next-best ``macro-action''. We borrow the notion of macro-action from planning under uncertainty~\citep{he2010puma} and define it as follows.
\begin{definition}[Macro-action]
 A macro-action is an exploration target (frontier) which is assumed to be reachable through an open-loop control strategy.
\end{definition}

\begin{figure}[t]
  \centering 
  \includegraphics[width=.6\columnwidth]{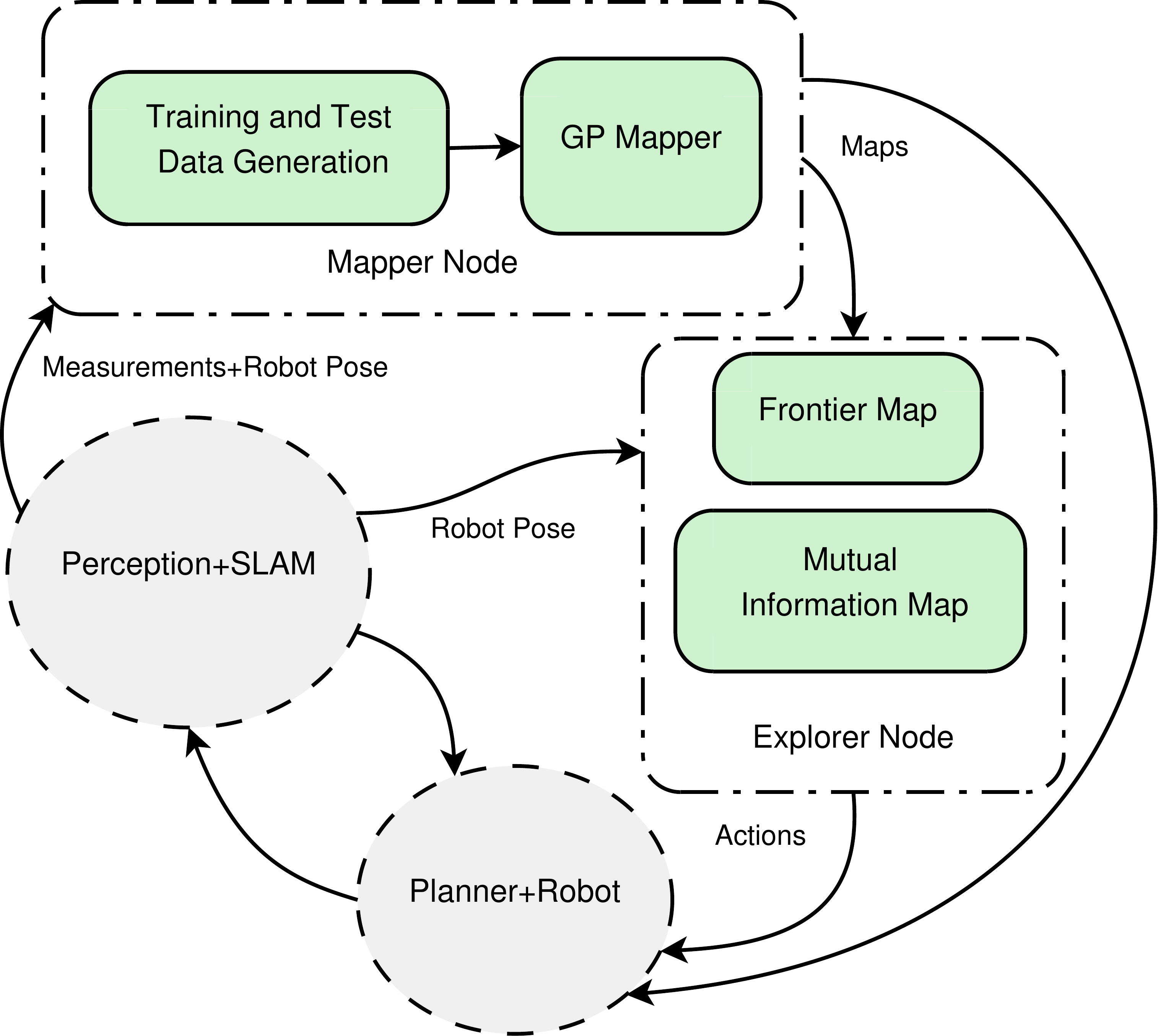}
  \caption{Schematic illustration of the autonomous mapping and exploration process using GPs maps. The GP mapper module provides the continuous occupancy map which can be exploited to extract geometric frontiers and mutual information maps. The maps also give support to the planner module for basic navigation tasks as well as cost-aware planning. The explorer node returns a macro-action (chosen frontier) that optimizes the expected utility function. The gray nodes are not investigated.}
  \label{fig:process}
\end{figure}

The employed measurement model is a standard beam-based mixture model for range-finder sensors~\citep{thrun2005probabilistic}, however, the proposed algorithm can be adapted to other sensor modalities with reasonable probabilistic observation models. Figure~\ref{fig:process} depicts the proposed mutual information-based navigation process concept using GPOMs.
\subsection{Contributions}
This article is based on our preliminary work~\citep{maani2015mi} where 
we presented continuous probabilistic frontier maps and an algorithm to calculate MI between the current map and 
future measurements. However, the distinguishable items from the previously published work are as follows:
\begin{itemize}
  \item We expand the decision making part and use the notion of macro-action.
  \item We provide details about the derivation of the sensor model and the construction of the training and test point sets.
  \item We present more detailed evaluations with comparable techniques.
\end{itemize}
The main contributions of this work are as follows.
\begin{itemize}
  \item A framework for incremental Gaussian processes occupancy mapping using range-finder sensors is developed. The method runs significantly faster with performance close to batch computation.
  \item We develop a novel probabilistic geometric frontier representation by exploiting continuous occupancy mapping and using $L^{1}$-norm of the gradient of continuous occupancy maps.
  \item A tractable technique for greedy information gain-based exploration is developed that takes into account all possible future observations. The technique can deal with sparse measurements and uses a forward sensor model for map predictions.
  \item The results from publicly available datasets in a highly structured indoor environment and a large-scale outdoor space are presented. 
\end{itemize}

\subsection{Notation}
In the present article probabilities and probability densities are not distinguished in general. Matrices are capitalized in bold, such as in $\boldsymbol X$, and vectors are in lower case bold type, such as in $\boldsymbol x$. Vectors are column-wise and $1\colon n$ means integers from $1$ to $n$. The Euclidean norm is shown by $\lVert \cdot \rVert$. $\lvert \boldsymbol X \rvert$ denotes the determinant of matrix $\boldsymbol X$. For the sake of compactness, random variables, such as $X$, and their realizations, $x$, are sometimes denoted interchangeably where it is evident from context. $x^{[i]}$ denotes a reference to the $i$-th element of the variable. An alphabet such as $\mathcal{X}$ denotes a set, and the cardinality of the set is denoted by $\lvert \mathcal{X} \rvert$. A subscript asterisk, such as in $\boldsymbol x_*$, indicates a reference to a test set quantity. The $n$-by-$n$ identity matrix is denoted by $\boldsymbol I_{n}$. $\mathrm{vec}(x^{[1]},\dots,x^{[n]})$ denotes a vector such as $\boldsymbol x$ constructed by stacking $x^{[i]}$, $\forall i \in \{1\colon n\}$. The function notation is overloaded based on the output type and denoted by $k(\cdot)$, $\boldsymbol k(\cdot)$, and $\boldsymbol K(\cdot)$ where the outputs are scalar, vector, and matrix, respectively. Finally, $\EV\cdot$, $\Var\cdot$, and $\Cov\cdot$ denote the expected value, variance, and covariance (for random vectors) of a random variable, respectively.

\subsection{Outline}
The remaining parts of this article are organized as follows. In Section~\ref{sec:literature}, a literature review is given. In Section~\ref{sec:mapping}, we present the proposed mapping algorithms and provide evaluations and comparison with occupancy grid maps. The exploration approach, MI surface calculation, and decision making process proposed in this work are discussed in Section~\ref{sec:Exploration}. Section~\ref{sec:Results} presents the results from experiments in two difference scenarios. Finally, Section~\ref{sec:conclusion} concludes the article and discusses the limitations of this work.

\section{Related work}
\label{sec:literature}
An environment can be explored by directing a robot towards frontiers that indicate unknown regions of the environment in the neighboring known free areas \citep{yamauchi1997frontier}. Traditional autonomous exploration strategies have been devised to use OGM~\citep{moravec1985high,elfes1987sonar,konolige1997improved,thrun2003learning,hornung2013octomap,merali2014optimizing} to represent free, occupied and unknown regions. The following works use the concept of \textit{active perception}~\citep{bajcsy1988active} to take actions that reduce the uncertainty in the state variables. A combined information utility for exploration is developed in \cite{bourgault2002information} using the information-based cost function in~\cite{feder1999adaptive} and an OGM. A one-step look-ahead strategy is used to generate the locally optimal control action. The reported results indicated that the utility for mapping attracts the robot to unknown areas while the localization utility keeps the robot well localized relative to known features in the map. In~\cite{makarenko2002experiment} an integrated exploration approach for a robot navigating in an unknown environment populated with beacons is proposed; a total utility function consisting of the weighted sum of the OGM entropy, navigation cost, and localizability is used. To enhance the map quality of the EKF-based Simultaneous Localization And Mapping (SLAM)~\citep{cadena2016past}, an \mbox{A-optimal} criterion for autonomous exploration is examined in~\cite{sim2005global}. Later in~\cite{carrillo2012comparison}, it is shown that the D-optimal (Determinant optimal) criterion~\citep{pukelsheim1993optimal} is more effective in such scenarios.

In~\cite{stachniss2005information}, Rao-Blackwellized Particle Filters (RBPF)~\citep{doucet2000rao} are employed to compute map and robot pose posteriors. The proposed uncertainty reduction approach is based on the joint entropy minimization of the SLAM posterior. The information gain is approximated using ray-casting for a given action. In~\cite{blanco2008novel}, through the entropy of the expected map of RBPF, the technique takes the uncertainty in both robot path and map into account. In a similar framework ~\cite{carlone2010application,carlone2014active} addressed the problem of active SLAM and exploration, specifically the inconsistency in the filter due to information loss for a given policy using the relative entropy concept. In~\cite{amigoni2010information}, it is assumed that all random variables are normally distributed and an exploration strategy based on relative entropy metric, combined traveling cost, and expected information gain is proposed.

The techniques in~\cite{valencia2012active,Vallve2013potential,Vallve2014dense,vallve2015potential} evaluate exploratory and place revisiting paths, which are selected based on entropy reduction estimates of both map and path. Similar to this work, these techniques use Pose SLAM~\citep{ila2010information, valencia2013planning}, a delayed-state SLAM algorithm from the pose graph family. Given the inherent complexity in the formulation to calculate the joint entropy of robot pose and map, it is assumed that they are conditionally independent. In~\cite{carrillo2015autonomous}, to avoid the need to update the map using unknown future measurements, the objective function is simplified to the current map entropy. In~\cite{kim2014active}, a greedy approach for active visual SLAM that considers area coverage and navigation uncertainty is proposed. In~\cite{Julian01092014}, the MI surface between a map and future measurements is computed numerically. The work assumes known robot poses, and relies on an OGM representation and measurements from a laser range-finder. The algorithm integrates over an information gain function with an inverse sensor model at its core. It is formally proven that any controller tasked to maximize an MI reward function is eventually attracted to unexplored areas. The technique in~\cite{charrow2015information} is closely related to this work. The computational performance of the information gain is increased by using Cauchy-Schwarz Quadratic Mutual Information (CSQMI). It is shown that the behavior of CSQMI is similar to that of MI while it can be computed faster. The technique has also been further extended to the multi-robot scenario~\citep{faigl2012goal,charrow2014approximate,charrow2015thesis}.

The methods reviewed above fall short of accounting for structural correlations in the environment. Kernel methods in the form of a Gaussian Processes framework~\citep{rasmussen2006gaussian} are non-parametric regression and classification techniques that have been extensively used by researchers to model spatial phenomena~\citep{lang2007adaptive,vasudevan2009gaussian,Hadsell01072010}. Gaussian Processes have proven particularly powerful to represent the affinity of spatially correlated data, hence overcoming the traditional assumption of independence between cells, characteristic of the occupancy grid method for mapping environments~\citep{o2009contextual,t2012gaussian}. The variance surface of GPs equate to a continuous representation of uncertainty in the environment, which it can be used to highlight unexplored regions and optimize a robot's search plan. The continuity property of the GP map can improve the flexibility of the planner by inferring directly on collected sensor data without being limited by the resolution of the grid cell~\citep{yang2013gaussian}. The incremental GP map building using the Bayesian Committee Machine (BCM) technique~\citep{tresp2000bayesian} is developed in~\cite{kim2012building,jadidi2013exploration,jadidi2013acra,maani2014com} and for online applications in~\cite{7487232}. In~\cite{ramoshilbert}, the Hilbert maps technique is proposed that is more scalable and can be updated in linear time. However, it approximates the problem and produces maps with less accuracy than GPOM.

GPOM, in its original formulation~\citep{o2009contextual,t2012gaussian}, is a batch mapping technique and the cubic time complexity of GPs (see Section~\ref{subsec:timecomplex}) is prohibitive for scenarios such as robotic navigation where a dense representation is preferred. The incremental GP map building was studied in~\citet{kim2012building}, and in~\citet{jadidi2013exploration,jadidi2013acra,maani2014com}.
In this work, we exploit GPs to develop tractable online robotic mapping and exploration techniques. We start from the problem of occupancy mapping and expand the method to exploration using geometric frontiers, and mutual information-based exploration.

\section{Mapping}
\label{sec:mapping}
\begin{figure}[t]
  \centering 
  \includegraphics[width=.7\columnwidth]{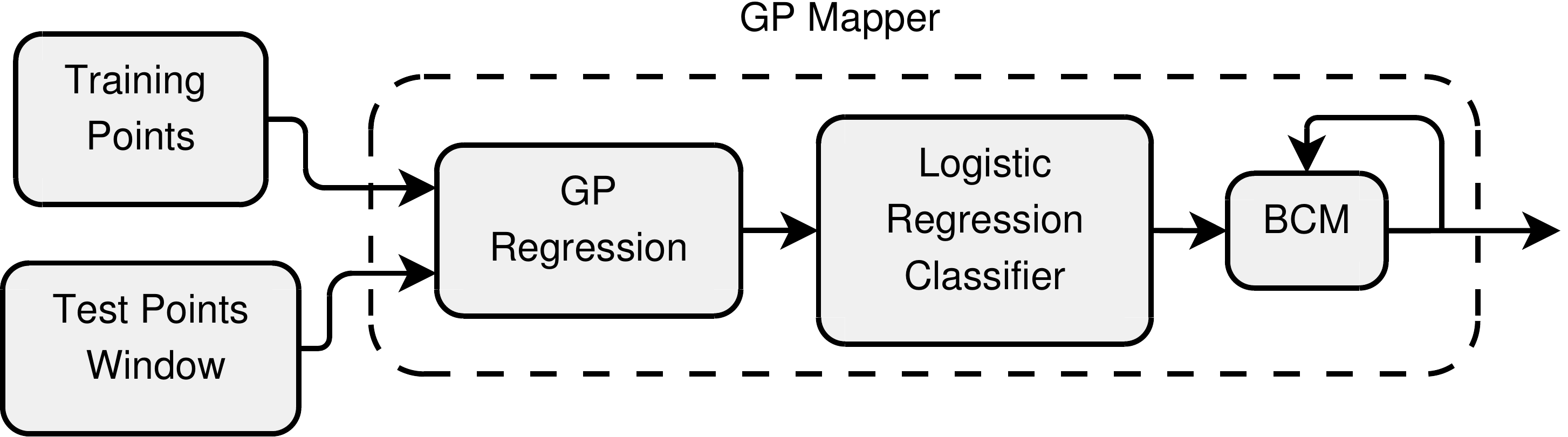}
  \caption{Schematic illustration of GP Mapper module. GP models the correlation in data and place distributions on test points. The logistic regression classifier squashes the output of GP into probabilities and returns the local map where the BCM module updates the global map incrementally.}
  \label{fig:mapper}
\end{figure}

The GP mapper module is shown in Figure~\ref{fig:mapper} which takes the processed measurements, i.e.\@ training data, and a test point window centered at the current robot pose as inputs to perform regression and classification steps for local maps generation and fuse them incrementally into the global frame through the BCM technique~\citep{tresp2000bayesian}.

Before formal statement of the problem, we clarify the following assumptions.
\begin{assumption}[Static environment]
The environment that the robot navigates in is static.
\end{assumption}
\begin{assumption}[Gaussian occupancy map points]
 Any sampled point from the occupancy map representation of the environment is a random variable whose distribution is Gaussian.
\end{assumption}

\subsection{Gaussian Processes}
\label{subsec:GPs}
A Gaussian Process is a collection of any finite number of random variables which are jointly distributed Gaussian \citep{rasmussen2006gaussian}. The joint distribution of the observed target values, $\boldsymbol y$, and the function values (the latent variable), $\boldsymbol f_*$, at the query points can be written as
\begin{equation}
\label{eq:gp_joint}
 \begin{bmatrix}
	\boldsymbol y \\
	\boldsymbol f_*
 \end{bmatrix} \sim \mathcal{N}(\boldsymbol 0,
 \begin{bmatrix}
	\boldsymbol K(\boldsymbol X,\boldsymbol X)+\sigma_n^2 \boldsymbol I_{n} & \boldsymbol K(\boldsymbol X,\boldsymbol X_*) \\
	\boldsymbol K(\boldsymbol X_*,\boldsymbol X)			& \boldsymbol K(\boldsymbol X_*,\boldsymbol X_*) 
 \end{bmatrix})
\end{equation}
where $\boldsymbol X$ is the $d\times n$ design matrix of aggregated input vectors $\boldsymbol x$, $\boldsymbol X_*$ is a $d\times n_*$ query points matrix, $\boldsymbol K(\cdot,\cdot)$ is the GP covariance matrix, and $\sigma_n^2$ is the variance of the observation noise which is assumed to have an independent and identically distributed (i.i.d.) Gaussian distribution. Define a training set \mbox{$\mathcal{D} = \{(\boldsymbol x^{[i]},y^{[i]}) \mid i=1\colon n\}$}. The predictive conditional distribution for a single query point $f_*|\mathcal{D},\boldsymbol x_* \sim \mathcal{N}(\EV{f_*},\Var{f_*})$ can be derived as 
\begin{equation}
 \label{eq:gp_mean}
 \mu = \EV{f_*} = \boldsymbol k(\boldsymbol X,\boldsymbol x_*)^{T}[\boldsymbol K(\boldsymbol X,\boldsymbol X)+\sigma_n^2 \boldsymbol I_{n}]^{-1}\boldsymbol y
\end{equation}
\begin{align}
\label{eq:gp_cov}
 \sigma = \Var{f_*} = k(\boldsymbol x_*,\boldsymbol x_*) - \boldsymbol k(\boldsymbol X,\boldsymbol x_*)^{T}[\boldsymbol K(\boldsymbol X,\boldsymbol X)+\sigma_n^2 \boldsymbol I_{n}]^{-1}\boldsymbol k(\boldsymbol X,\boldsymbol x_*)
\end{align}

The Mat\'ern family of covariance functions \citep{stein1999interpolation} has proven powerful features to model structural correlations \citep{jadidi2013exploration,kim2013occupancy,maani2014com,kim2015gpmap} and hereby we select them as the kernel of GPs. For a single query point $\boldsymbol x_*$ the function is given by 
\begin{align}
\label{eq:Matern}
k(\boldsymbol x,\boldsymbol x_*) = \frac{1}{\Gamma(\nu) 2^{\nu-1}}\left[\frac{\sqrt{2\nu}\lVert \boldsymbol x - \boldsymbol x_* \rVert}{\kappa}\right]^{\nu} K_{\nu}\left(\frac{\sqrt{2\nu}\lVert \boldsymbol x - \boldsymbol x_* \rVert}{\kappa} \right)
\end{align}
where $\Gamma$ is the Gamma function, $K_{\nu}(\cdot)$ is the modified Bessel function of the second kind of order $\nu$, $\kappa$ is the characteristic length scale, and $\nu$ is a positive parameter used to control the smoothness of the covariance.

The hyperparameters of the covariance and mean function, $\boldsymbol\theta$, can be computed by minimization of the negative log of the marginal likelihood (NLML) function.
\begin{align}
\label{eq:nlml}
	\log p(\boldsymbol y|\boldsymbol X,&\boldsymbol\theta) = -\frac{1}{2}\boldsymbol y^{T}[\boldsymbol K(\boldsymbol X,\boldsymbol X)+\sigma_n^2 \boldsymbol I_{n}]^{-1}\boldsymbol y -\frac{1}{2}\log \arrowvert K(\boldsymbol X,\boldsymbol X)+\sigma_n^2 \boldsymbol I_{n} \arrowvert-\frac{n}{2}\log 2\pi\
\end{align}

\subsection{Problem statement and formulation}
\label{subsec:statement}
Let $\mathcal{M}$ be the set of possible occupancy maps. We consider the map of the environment to be static and as an $n_m$-tuple random variable \mbox{$(M^{[1]},...,M^{[n_m]})$} whose elements are described by a normal distribution \mbox{$m^{[i]} \sim \mathcal{N}(\mu^{[i]},\sigma^{[i]})$, $i \in \{1\colon n_m\}$}. Let \mbox{$\mathcal{Z} \subset \mathbb{R}_{\geq 0}$} be the set of possible range measurements. The observation consists of an $n_z$-tuple random variable $(Z^{[1]},...,Z^{[n_z]})$ whose elements can take values \mbox{$\boldsymbol z^{[k]} \in \mathcal{Z}$, $k \in \{1\colon n_z\}$}. Let $\mathcal{X} \subset \mathbb{R}^2$ be the set of spatial coordinates to build a map on. Let $\boldsymbol x_o^{[k]} \in \mathcal{X}_o \subset \mathcal{X}$ be an observed occupied point by the $k$-th sensor beam from the environment which, at any time-step $t$, can be calculated by transforming the local observation $\boldsymbol z^{[k]}$ to the global frame using the robot pose $\boldsymbol x_t \in \mathrm{SE(2)}$. Let $\boldsymbol X_f^{[k]} \in \mathcal{X}_f \subset \mathcal{X}$ be the matrix of sampled unoccupied points from a line segment with the robot pose and corresponding observed occupied point as its endpoints. Let $\mathcal{D}=\mathcal{D}_o \cup \mathcal{D}_f$ be the set of all training points. We define a training set of occupied points \mbox{$\mathcal{D}_o = \{(\boldsymbol x_o^{[i]},y_o^{[i]}) \mid i=1\colon n_o\}$} and a training set of unoccupied points \mbox{$\mathcal{D}_f = \{(\boldsymbol x_f^{[i]},y_f^{[i]}) \mid i=1\colon n_f\}$} in which  \mbox{$\boldsymbol y_o = \mathrm{vec}(y_o^{[1]},...,y_o^{[n_o]})$} and \mbox{$\boldsymbol y_f = \mathrm{vec}(y_f^{[1]},...,y_f^{[n_f]})$} are target vectors and each of their elements can belong to the set $\mathcal{Y}=\{-1,+1\}$ where $-1$ and $+1$ corresponds to unoccupied and occupied locations, respectively,  $n_o$ is the total number of occupied points, and $n_f$ is the total number of unoccupied points. Given the robot pose $\boldsymbol x_t$ and observations $Z_t= \boldsymbol z_t$, we wish to estimate \mbox{$p(M=m\mid \boldsymbol x_t, Z_t= \boldsymbol z_t)$}.  Place a joint distribution over $M$; the map can be inferred as a Gaussian process by defining the process as the function $y:\mathcal{X}\rightarrow\mathcal{M}$, therefore
\begin{equation}
 \label{eq:mapGP}
 y(\boldsymbol x) \sim \mathcal{GP}(f_m(\boldsymbol x), k(\boldsymbol x,\boldsymbol x'))
\end{equation}
It is often the case that we set the mean function $f_m(\boldsymbol x)$ as zero, unless it is mentioned explicitly that $f_m(\boldsymbol x)\neq0$. For a given query point in the map, $\boldsymbol x_*$, GP predicts a mean, $\mu$, and an associated variance, $\sigma$. We can write
\begin{equation}
 \label{eq:mapy}
 m^{[i]} = y(\boldsymbol x^{[i]}_*) \sim \mathcal{N}(\mu^{[i]},\sigma^{[i]})
\end{equation}
To show a valid probabilistic representation of the map $p(m^{[i]})$, the classification step, a logistic regression classifier~\citep[Sections 3.1 and 3.2]{rasmussen2006gaussian},~\citep[Chapter 8]{murphy2012machine},~\citep{maani2014com}, squashes data into the range $[0,1]$.

\subsection{Sensor model, training and test data}
\label{subsec:sensor}
\begin{figure}[t]
  \centering 
  \includegraphics[width=.45\columnwidth]{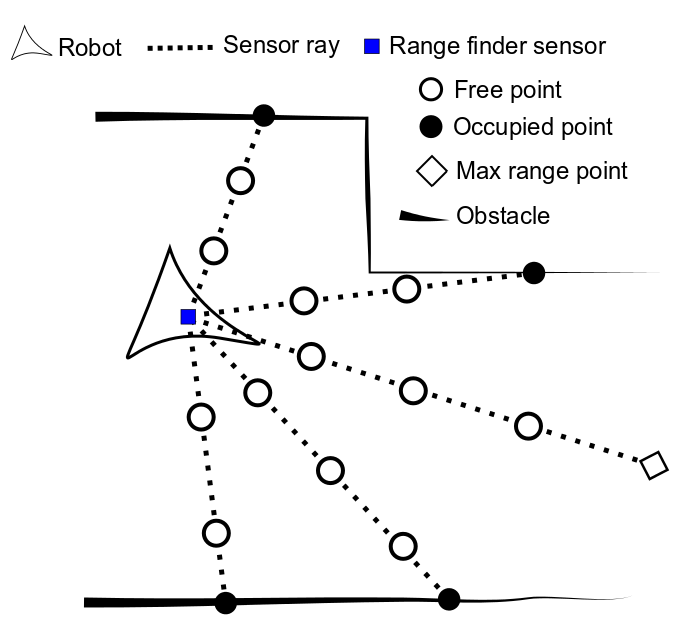}
  \caption{Conceptual illustration of the robot, the environment, and observations. Training data consists of free and occupied points labeled $y_f=-1$ and $y_o=+1$ respectively. Free points are sampled along each beam, i.e.\@ negative sensor information while occupied points are directly observable.}
  \label{fig:setup}
\end{figure}

The robot is assumed to be equipped with a 2D range-finder sensor. The raw measurements include points returned from obstacle locations. For any sensor beam, the distance from the sensor position to the detected obstacle along that beam indicates a line from the unoccupied region of the environment. To build training data points for the unoccupied part of the map, it is required to sample along the aforementioned line. Figure~\ref{fig:setup} shows the conceptual illustration of the environment and training points generation.

A sensor beam $\boldsymbol z_t = (\boldsymbol z_t^{[1]},...,\boldsymbol z_t^{[n_z]})$ has $n_z$ range observations at a specific bearing depending on the density of the beam. The observation model for each $\boldsymbol z_t^{[k]}$ can be written as
\begin{equation}
 \boldsymbol z_t^{[k]} =
 \begin{bmatrix}
	r_t^{[k]} \\
	\alpha_t^{[k]}
 \end{bmatrix} = h(\boldsymbol x_t,\boldsymbol x_o^{[k]})+\boldsymbol v, \quad \boldsymbol v \sim \mathcal{N}(\boldsymbol 0,\boldsymbol R)
\end{equation}
\begin{equation}
  h(\boldsymbol x_t,\boldsymbol x_o^{[k]}) \triangleq 
  \begin{bmatrix}
	\sqrt{(\boldsymbol x_o^{[k,1]} - \boldsymbol x_t^{[1]})^2 + (\boldsymbol x_o^{[k,2]} - \boldsymbol x_t^{[2]})^2}\\
	\arctan(\boldsymbol x_o^{[k,2]} - \boldsymbol x_t^{[2]},\boldsymbol x_o^{[k,1]} - \boldsymbol x_t^{[1]}) - \boldsymbol x_t^{[3]}
 \end{bmatrix}
\end{equation}
where $r_t^{[k]}$ is the range measurement from the $k$-th sensor beam and $\alpha_t^{[k]}$ is the corresponding angle of $r_t^{[k]}$. 
The observation model noise $\boldsymbol v$ is assumed to be Gaussian with zero mean and covariance $\boldsymbol R$. To find $\boldsymbol x_o^{[k]}$ which is in the map space, the inverse model can be calculated as
\begin{equation}
\label{eq:occnt}
  \boldsymbol x_o^{[k]} = \boldsymbol x_t^{[1:2]} + r_t^{[k]} R(\boldsymbol x_t^{[3]}) 
 \begin{bmatrix}
	\cos(\alpha_t^{[k]}) \\
	\sin(\alpha_t^{[k]})
 \end{bmatrix}
\end{equation}
where $R(\boldsymbol x_t^{[3]}) \in \mathrm{SO(2)}$ indicates a $2\times2$ rotation matrix.

Having defined the observed occupied points in the map space, now we can construct the training set of occupied points as \mbox{$\mathcal{D}_o = \{(\boldsymbol x_o^{[k]},y_o^{[k]}) \mid k=1\colon n_z\}$}. 
One simple way to build the free area training points is to uniformly sample along the line segment, $l_z^{[k]}$, with the robot position and any occupied point $\boldsymbol x_o^{[k]}$ as its end points. Therefore,
\begin{equation}
\label{eq:unoccnt}
 \boldsymbol X_f^{[k,j]} = \boldsymbol x_t^{[1:2]} + \delta_j R(\boldsymbol x_t^{[3]})
 \begin{bmatrix}
	\cos(\alpha_t^{[k]}) \\
	\sin(\alpha_t^{[k]})
 \end{bmatrix}
\end{equation}
where \mbox{$\delta_j \sim \mathcal{U}(0, r_t^{[k]}) \quad j = 1\colon n_f^{[k]}$}, $\mathcal{U}(0, r_t^{[k]})$ is a uniform distribution with the support $[0,r_t^{[k]}]$ and $n_f^{[k]}$ is the desired number of samples for the $k$-th sensor beam. $n_f^{[k]}$ can be a fixed value for all the beams or variable, e.g.\@ a function of the line segment length $\lVert l_z^{[k]}\rVert=r_t^{[k]}$. In the case of a variable number of points for each beam, it is useful to set a minimum value $n_{fmin}$. Therefore we can write
\begin{equation}
 n_f^{[k]} \triangleq \max (\{n_{fmin},s_l(r_t^{[k]})\})
\end{equation}
where $s_l(\cdot)$ is a function that adaptively generates a number of sampled points based on the input distance. This minimum value controls the sparsity of the training set of unoccupied points. Alternatively, we can select a number of equidistant points instead of sampling. However, as the number of training points increases, the computational time grows cubicly. We can construct the training set of unoccupied points as \mbox{$\mathcal{D}_f = \bigcup_{i=1}^{n_z} \mathcal{D}_f^{[i]}$} where \mbox{$\mathcal{D}_f^{[i]} = \{(\boldsymbol X_f^{[k]},\boldsymbol y_f^{[k]}) \mid k=1\colon n_z\}$} and \mbox{$\boldsymbol y_f^{[k]} = \mathrm{vec}(y_f^{[1]},...,y_f^{[n_f^{[k]}]})$}.

\begin{remark}
Generally speaking, query points can have any desired distributions and the actual representation of the map depends on that distribution. However, building the map over a grid facilitates comparison with standard occupancy grid-based methods, i.e.\@ at similar map resolutions. We use function \texttt{TestDataWindow}, in Algorithms~\ref{alg:GPOM} and \ref{alg:GPOM2}, for generating a grid at a given position. The size of this grid can be set according to the maximum sensor range, the environment size, or available computational resources for data processing.
\end{remark}

\begin{remark}
Throughout all algorithms, when we write $m$ for a map, it is assumed that the mean $\boldsymbol\mu$, the variance $\boldsymbol\sigma$, the occupancy probability $p(m)$, and the corresponding spatial coordinates are available even if they are not mentioned or used explicitly. For simplicity, when $m$ is used for computations such as in $\log(p(m))$, we write $\log(m)$.
\end{remark}

\subsection{Map management}
\label{subsec:management}

An important advantage of a mapping method is its capability to use past information appropriately. The mapping module returns local maps centered at the robot pose. Therefore, in order to keep track of the global map, a map management step is required where the local inferred map can be fused with the current global map. This incremental approach allows for handling larger map sizes, and map inference at the local level is independent of the global map.

To incorporate new information incrementally, map updates are performed using BCM. The technique combines estimators which were trained on different data sets. Assuming a Gaussian prior with zero mean and covariance $\boldsymbol\Sigma$ and each GP with mean $\EV{f_*|\mathcal{D}^{[i]}}$ and covariance $\Cov{f_*|\mathcal{D}^{[i]}}$, it follows that~\citep{tresp2000bayesian}
\vspace{-0.1cm}
\begin{equation}	
\label{eq:bcm}	
	\EV{f_*|\mathcal{D}} = \boldsymbol{C}^{-1} \sum_{i=1}^{p_m}\Cov{f_*|\mathcal{D}^{[i]}}^{-1}\EV{f_*|\mathcal{D}^{[i]}}
\end{equation}
\begin{equation}
	\label{eq:bcm2}	
	 \boldsymbol{C} = \Cov{f_*|\mathcal{D}}^{-1} = -(p_m-1)(\boldsymbol\Sigma)^{-1} + \sum_{i=1}^{p_m}\Cov{f_*|\mathcal{D}^{[i]}}^{-1}
\end{equation}
where $p_m$ is the total number of mapping processes. In this work, we use BCM for combining a local and a previously existing global map, or merging two global maps; therefore $p_m = 2$. In addition, in the case of uninformative prior over map points the term $\boldsymbol\Sigma^{-1}$ can be set to zero, i.e.\@ very large covariances/variances.

\begin{algorithm}[t!]
\caption[IGPOM]{\texttt{IGPOM}()}
\label{alg:GPOM}
\begin{algorithmic}[1]
\Require Robot pose $\boldsymbol p$ and measurements $\boldsymbol z$;
\If{$\mathrm{firstFrame}$}
\State $m \gets \varnothing$ $\quad$ // Initialize the map
\State optimize GP hyperparameters $\boldsymbol\theta$ // Minimize the NLML, Equation~\eqref{eq:nlml}
\EndIf
\State $\boldsymbol X_* \gets \texttt{TestDataWindow}(\boldsymbol p)$ // Query points grid centered at the robot pose
\State $\boldsymbol X_o, \boldsymbol y_o \gets \texttt{Transform2Global}(\boldsymbol p, \boldsymbol z)$ // Occupied training data, label $+1$, Equation~\eqref{eq:occnt}
\State $\boldsymbol X_f, \boldsymbol y_f \gets \texttt{TrainingData}(\boldsymbol p, \boldsymbol z)$ // Unoccupied training data, label $-1$, Equation~\eqref{eq:unoccnt}
\State $[\boldsymbol\mu_*, \boldsymbol\sigma_*] \gets \texttt{GP}(\boldsymbol\theta, [\boldsymbol X_o; \boldsymbol X_f], [\boldsymbol y_o; \boldsymbol y_f], \boldsymbol X_*)$ // Compute predictive mean and variance, Equation~\eqref{eq:gp_mean} and \eqref{eq:gp_cov}
\State $m \gets \texttt{UpdateMap}(\boldsymbol\mu_*,\boldsymbol\sigma_*, m)$ // Algorithm~\ref{alg:update}
\Return $m$
\end{algorithmic}
\end{algorithm}

\begin{algorithm}[t]
\caption[FusionBCM]{\texttt{FusionBCM}($\mu_a, \mu_b, \sigma_a, \sigma_b$)}
\label{alg:bcm}
\begin{algorithmic}[1]
\State $\sigma_c \gets (\sigma_a^{-1}+\sigma_b^{-1})^{-1}$ // Point-wise calculation of Equation~\eqref{eq:bcm2}
\State $\mu_c \gets \sigma_c(\sigma_a^{-1}\mu_a + \sigma_b^{-1}\mu_b)$ // Point-wise calculation of Equation~\eqref{eq:bcm}
\Return $\mu_c, \sigma_c$
\end{algorithmic}
\end{algorithm}

\begin{algorithm}[t]
\caption[UpdateMap]{\texttt{UpdateMap}()}
\label{alg:update}
\begin{algorithmic}[1]
\Require Global map $m$, $\boldsymbol\mu$, $\boldsymbol\sigma$ and local map $m_*$, $\boldsymbol\mu_*$, $\boldsymbol\sigma_*$;
\For {all $i\in\mathcal{M}_*$}
\State $j \gets$ find the corresponding global index of $i$ using the map spatial coordinates and a nearest neighbor search
\State $\boldsymbol\mu^{[j]}, \boldsymbol\sigma^{[j]} \gets \texttt{FusionBCM}(\boldsymbol\mu^{[j]}, \boldsymbol\mu_*^{[i]}, \boldsymbol\sigma^{[j]}, \boldsymbol\sigma_*^{[i]})$ // Algorithm~\ref{alg:bcm}
\EndFor
\State $m \gets \texttt{LogisticRegression}(\boldsymbol\mu, \boldsymbol\sigma)$ // Squash data into (0,1)
\Return $m$
\end{algorithmic}
\end{algorithm}

\begin{figure}[t]
  \centering 
  \subfloat{\includegraphics[width=.5\columnwidth,trim={0.5cm 0cm 2cm 0cm},clip]{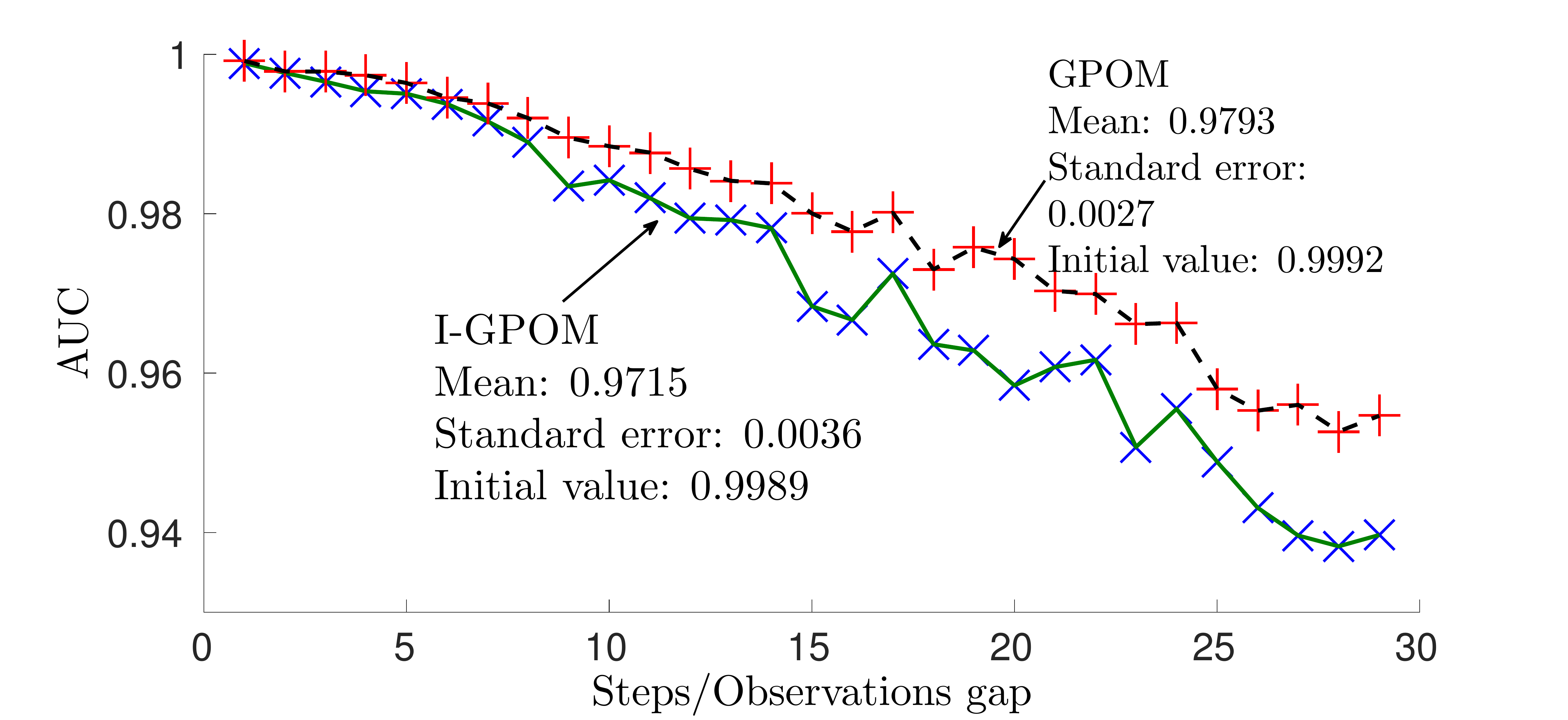}
  \label{fig:bcmeffect}}
  \subfloat{\includegraphics[width=.5\columnwidth,trim={0.5cm 0cm 2cm 0cm},clip]{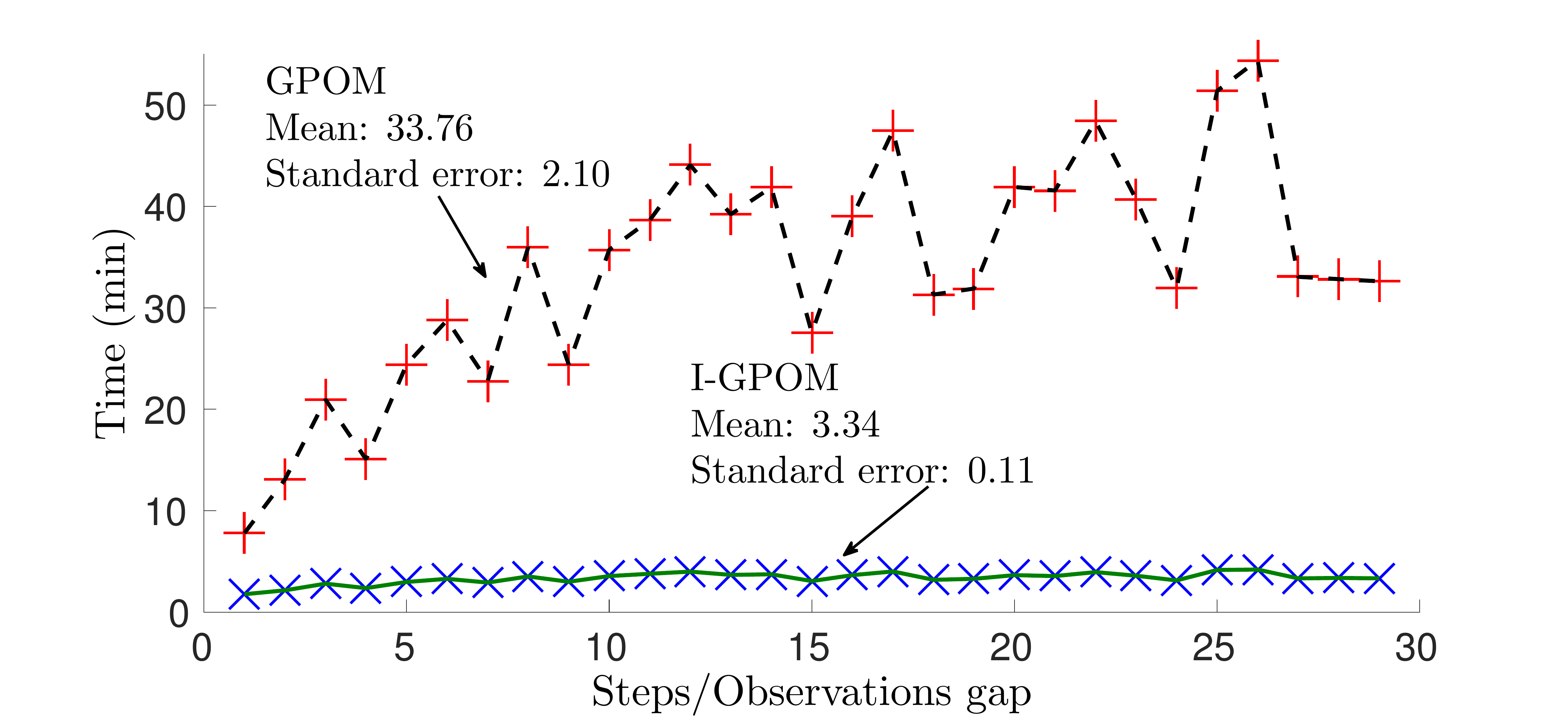}
  \label{fig:bcmeffecttime}} 
  \caption{Comparison of I-GPOM and batch GPOM methods using the Intel dataset with the observations size of $25$ laser scans at each step due to the memory limitation for the batch GP computations. The left plot shows the AUC and the right plot depicts the runtime for each step. The horizontal axes indicate observations gaps. As the number of gaps grows, the batch GP outperforms the incremental method as it learns the correlation between observations at once; however, with higher computational time. On the other hand, the incremental method in nearly constant time per update produces a similar average map quality with the mean difference of $0.0078$.}
  \label{fig:bcmanalysis}
\end{figure}

The steps of the incremental GPOM (I-GPOM) are shown in Figure~\ref{fig:mapper} and Algorithms~\ref{alg:GPOM}, \ref{alg:bcm}, and \ref{alg:update}, where a BCM module updates the global map as new observations are taken. In Figure~\ref{fig:bcmanalysis} a comparison of the incremental (\mbox{I-GPOM}) and batch (GPOM) GP occupancy mapping using the Intel dataset~\citep{Radish_data_set} with respect to the area under the receiver operating characteristic curve (AUC) and runtime is presented. The probability that the classifier ranks a randomly chosen positive instance higher than a randomly chosen negative instance can be understood using the AUC of the classifier; furthermore, the AUC is useful for domains with skewed class distribution and unequal classification error costs~\citep{fawcett2006introduction}. Without loss of generality, a set of $25$ laser scans, where each scan contains about $180$ points, had to be set due to the memory limitation imposed by the batch GP computations with a growing gap between successive laser scans from $1$ to $29$. The proposed incremental mapping approach using BCM performs accurate and close to the batch form even with about $8$ steps intermission between successive observations and is faster.

Optimization of the hyper-parameters is performed once at the beginning of each experiment by minimization of the negative log of the marginal likelihood function. For the prevailing case of multiple runs in the same environment, the optimized values can then be loaded off-line.

\subsection{I-GPOM2; an improved mapping strategy}
\label{challanges}

Inferring a high quality map compatible with the actual shape of the environment
can be non-trivial (see Figure~9 in \citet{t2012gaussian} and Figure~3 in \citet{kim2013continuous}). 
Although considering correlations of map points through regression 
results in handling sparse measurements, training a unique GP for both
occupied and free areas has two major challenges:
\begin{itemize}
 \item It limits the selection of an appropriate kernel that suits both occupied and unoccupied regions of the map,  
  effectively resulting in poorly extrapolated obstacles or low quality free areas.
 \item Most importantly, it leads to a mixed variance surface. In other words, it is not
  possible to disambiguate between boundaries of occupied-unknown and
  free-unknown space, unless the continuous map is thresholded (see Figure~6 in \citet{t2012gaussian}).
\end{itemize}

\begin{figure}[t!]
  \centering 
  \subfloat{\includegraphics[width=.3\columnwidth]{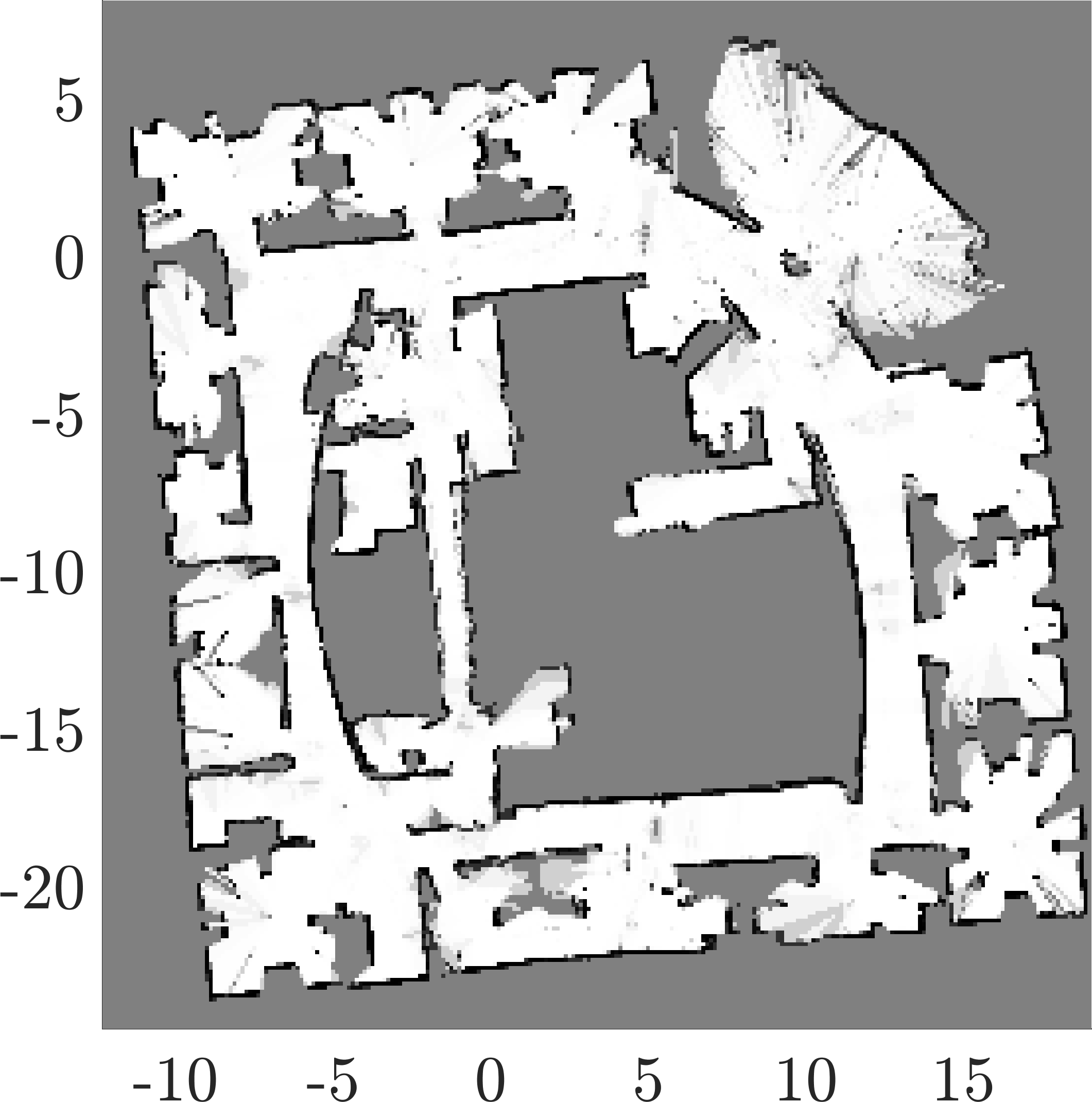}\label{fig:ogmap}}~
  \subfloat{\includegraphics[width=.345\columnwidth]{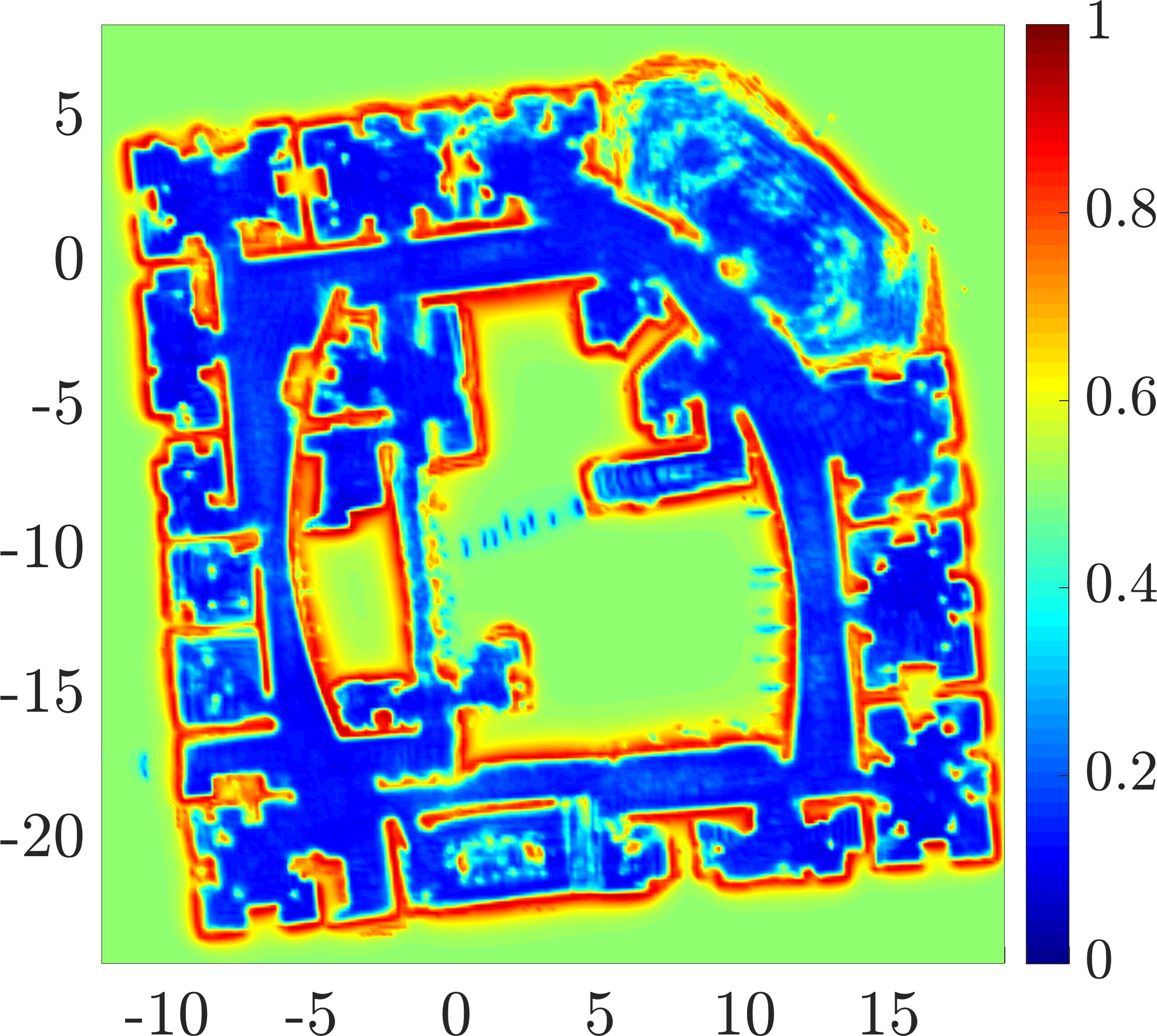}\label{fig:gpmap}}~
  \subfloat{\includegraphics[width=.345\columnwidth]{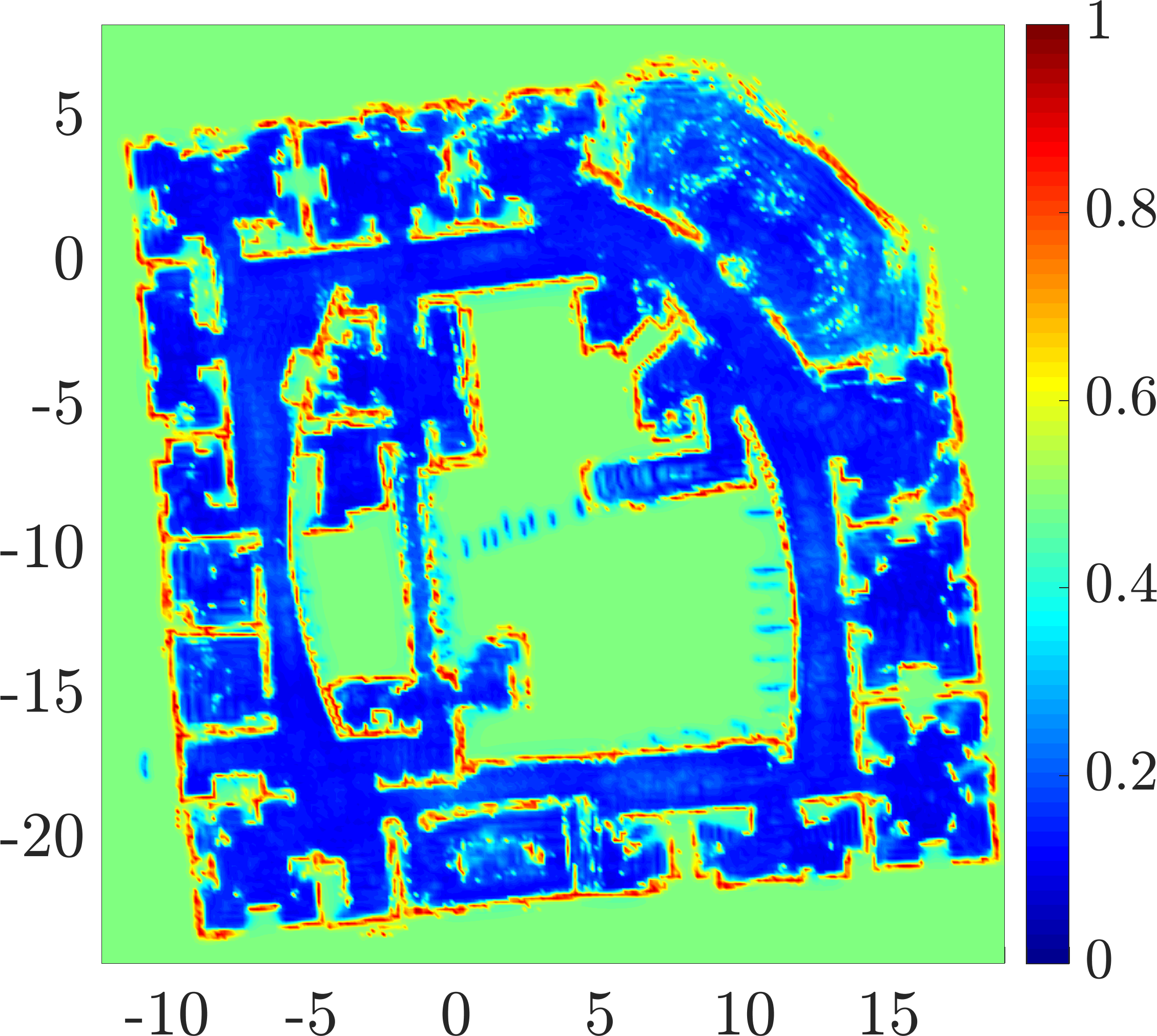}\label{fig:gpmap2}}
  \caption{Occupancy maps visualization; from left to right: OGM, I-GPOM, I-GPOM2. The maps are build incrementally using all observations available in the Intel dataset. For the I-GPOM and I-GPOM2 maps the Mat\'ern ($\nu = 3/2$) covariance function is used. I-GPOM and I-GPOM2 can complete partially observable areas, i.e.\@ incomplete areas in the OGM; however, using two GP in I-GPOM2 method produces more accurate maps for navigation purposes. The SLAM problem is solved by using the Pose SLAM algorithm and the map qualities depend on the robot localization accuracy.}
  \label{fig:Intelmaps}
\end{figure}

The first problem is directly related to the inferred map quality, while the second is a challenge for exploration using continuous occupancy maps. The integral kernel approach \citep{o2011continuous} can mitigate the first aforementioned deficiency, however, the integration over GPs kernels is computationally demanding and results in less tractable methods. In order to address these problems we propose training two separate GPs, one for free areas and one for obstacles, and merge them to build a unique continuous occupancy map (I-GPOM2). The complete results of occupancy mapping with the three different methods in the Intel dataset are presented in Figure~\ref{fig:Intelmaps}, while the AUCs are compared in Table \ref{tab:aucroc}. The I-GPOM2 method demonstrates more flexibility to model the cluttered rooms and has higher performance than the other methods. The ground truth map was generated using the registered points map and an image dilation technique to remove outliers. In this way, the ground truth map has the same orientation which makes the comparison convenient. GPOM-based maps infer partially observed regions; however, in the absence of a complete ground truth map, this fact can be only verified using Figure~\ref{fig:Intelmaps} and is not reflected in the AUC of I-GPOM and I-GPOM2. Algorithms~\ref{alg:GPOM2} and \ref{alg:merge} encapsulate the I-GPOM2 methods as implemented in the present work.

\begin{table}[!t]
\footnotesize
\centering
\caption{Comparison of the AUC and runtime for OGM, I-GPOM, and I-GPOM2 using the Intel dataset.}
\begin{tabular}{lcc}
\toprule
Method			& AUC		& Runtime (min) \\ \midrule

OGM			& 0.9300	& 7.28 	\\
I-GPOM			& 0.9439	& 102.44 	\\
I-GPOM2			& 0.9668	& 114.53 \\ \bottomrule
\end{tabular}
\label{tab:aucroc}

\end{table}

\subsection{Frontier map}
\label{subsec:Frontier_Maps}
Constructing a frontier map is the fundamental ingredient of any geometry-based exploration approach. It reveals the boundaries between known-free and unknown areas which are potentially informative regions for map expansion. In contrast to the classical binary representation, defining frontiers in a probabilistic form using map uncertainty is more suitable for computing expected behaviors. The boundaries that correspond to frontiers can be computed using the following heuristic formula.
\begin{equation}
\label{eq:frontier}
	\bar{f}^{[i]} \triangleq \lVert\nabla p(m^{[i]})\rVert_1 - \beta(\lVert\nabla p(m_o^{[i]})\rVert_1 + p(m_o^{[i]}) - 0.5)
\end{equation}
where $\nabla$ denotes the gradient operator, and $\beta$ is a factor that controls the effect of obstacle boundaries. $\lVert\nabla p(m^{[i]})\rVert_1$ indicates all boundaries while $\lVert\nabla p(m_o^{[i]})\rVert_1$ defines obstacle outlines. The subtracted constant is to remove the biased probability for unknown areas in the obstacles probability map.

\begin{algorithm}[t]
\caption[IGPOM2]{\texttt{IGPOM2}()}
\label{alg:GPOM2}
\begin{algorithmic}[1]
\Require Robot pose $\boldsymbol p$ and measurements $\boldsymbol z$;
\If{$\mathrm{firstFrame}$}
\State $m, m_o, m_f \gets \varnothing$ $\quad$ // Initialize the map
\State optimize GP hyperparameters $\boldsymbol\theta_o$, $\boldsymbol\theta_f$ // Minimize the NLML, Equation~\eqref{eq:nlml}
\EndIf
\State $\boldsymbol X_* \gets \texttt{TestDataWindow}(\boldsymbol p)$ // Query points grid centered at the robot pose
\State $\boldsymbol X_o, \boldsymbol y_o \gets \texttt{Transform2Global}(\boldsymbol p, \boldsymbol z)$ // Occupied training data, label $+1$, Equation~\eqref{eq:occnt}
\State $\boldsymbol X_f, \boldsymbol y_f \gets \texttt{TrainingData}(\boldsymbol p, \boldsymbol z)$ // Unoccupied training data, label $-1$, Equation~\eqref{eq:unoccnt}
\State $[\boldsymbol\mu_{o*}, \boldsymbol\sigma_{o*}] \gets \texttt{GP}(\boldsymbol\theta_o, \boldsymbol X_o, \boldsymbol y_o, \boldsymbol X_*)$ // Compute occupied map predictive mean and variance, Equation~\eqref{eq:gp_mean} and \eqref{eq:gp_cov}
\State $[\boldsymbol\mu_{f*}, \boldsymbol\sigma_{f*}] \gets \texttt{GP}(\boldsymbol\theta_f, \boldsymbol X_f, \boldsymbol y_f, \boldsymbol X_*)$ // Compute unoccupied map predictive mean and variance using \eqref{eq:gp_mean} and \eqref{eq:gp_cov}
\State $m_o \gets \texttt{UpdateMap}(\boldsymbol\mu_{o*}, \boldsymbol\sigma_{o*}, m_o)$ // Algorithm~\ref{alg:update}
\State $m_f \gets \texttt{UpdateMap}(\boldsymbol\mu_{f*}, \boldsymbol\sigma_{f*}, m_f)$ 
\State $m \gets \texttt{MergeMap}(m_o, m_f)$ // Algorithm~\ref{alg:merge}
\Return $m, m_o$
\end{algorithmic}
\end{algorithm}

\begin{algorithm}[t!]
\caption[MergeMap]{\texttt{MergeMap}()}
\label{alg:merge}
\begin{algorithmic}[1]
\Require Unoccupied map $m_f$, $\boldsymbol\mu_f$, $\boldsymbol\sigma_f$ and occupied map $m_o$, $\boldsymbol\mu_o$, $\boldsymbol\sigma_o$; 
\For {all $i\in\mathcal{M}$}
\State $\boldsymbol\mu^{[i]}, \boldsymbol\sigma^{[i]} \gets \texttt{FusionBCM}(\boldsymbol\mu_{o}^{[i]}, \boldsymbol\mu_{f}^{[i]}, \boldsymbol\sigma_{o}^{[i]}, \boldsymbol\sigma_{f}^{[i]})$ // Algorithm~\ref{alg:bcm}
\EndFor
\State $m \gets \texttt{LogisticRegression}(\boldsymbol\mu, \boldsymbol\sigma)$ // Squash data into (0,1)
\Return $m$
\end{algorithmic}
\end{algorithm}

\begin{figure}[!t]
  \centering 
  \subfloat{\includegraphics[width=.32\columnwidth]{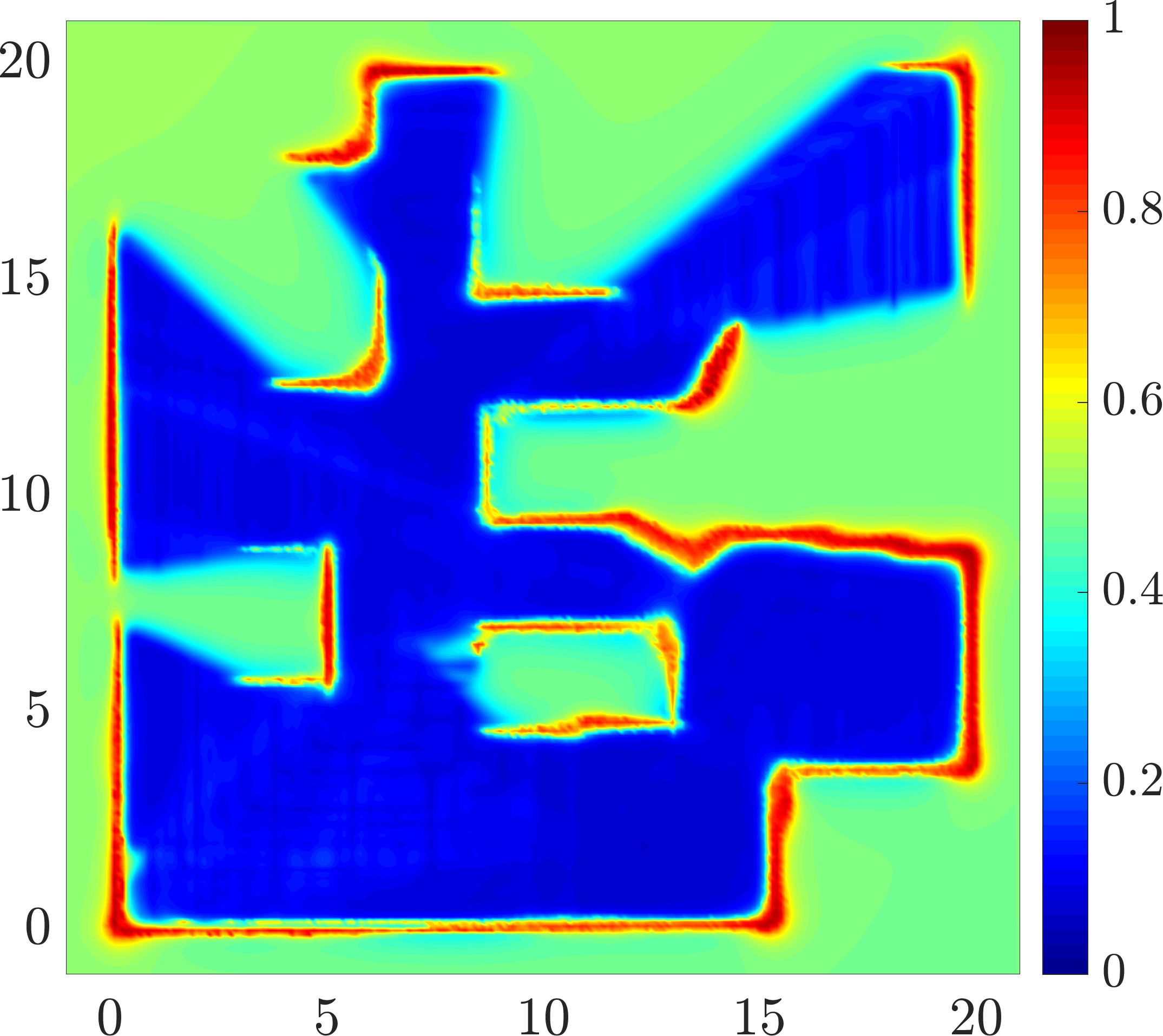}
  \label{fig:ex_com}}~
  \subfloat{\includegraphics[width=.32\columnwidth]{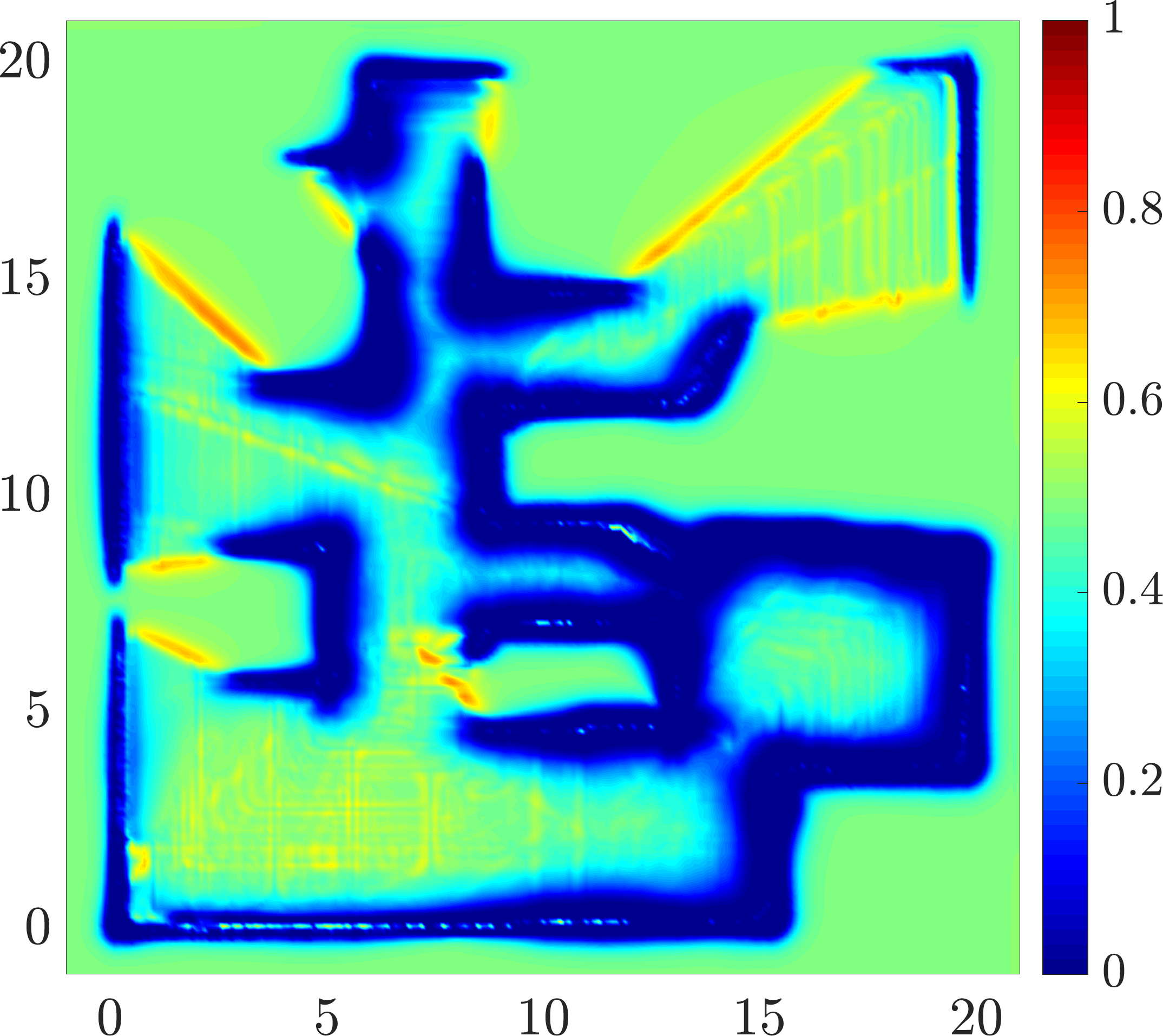}
  \label{fig:frontier}}~
  \subfloat{\includegraphics[width=.32\columnwidth]{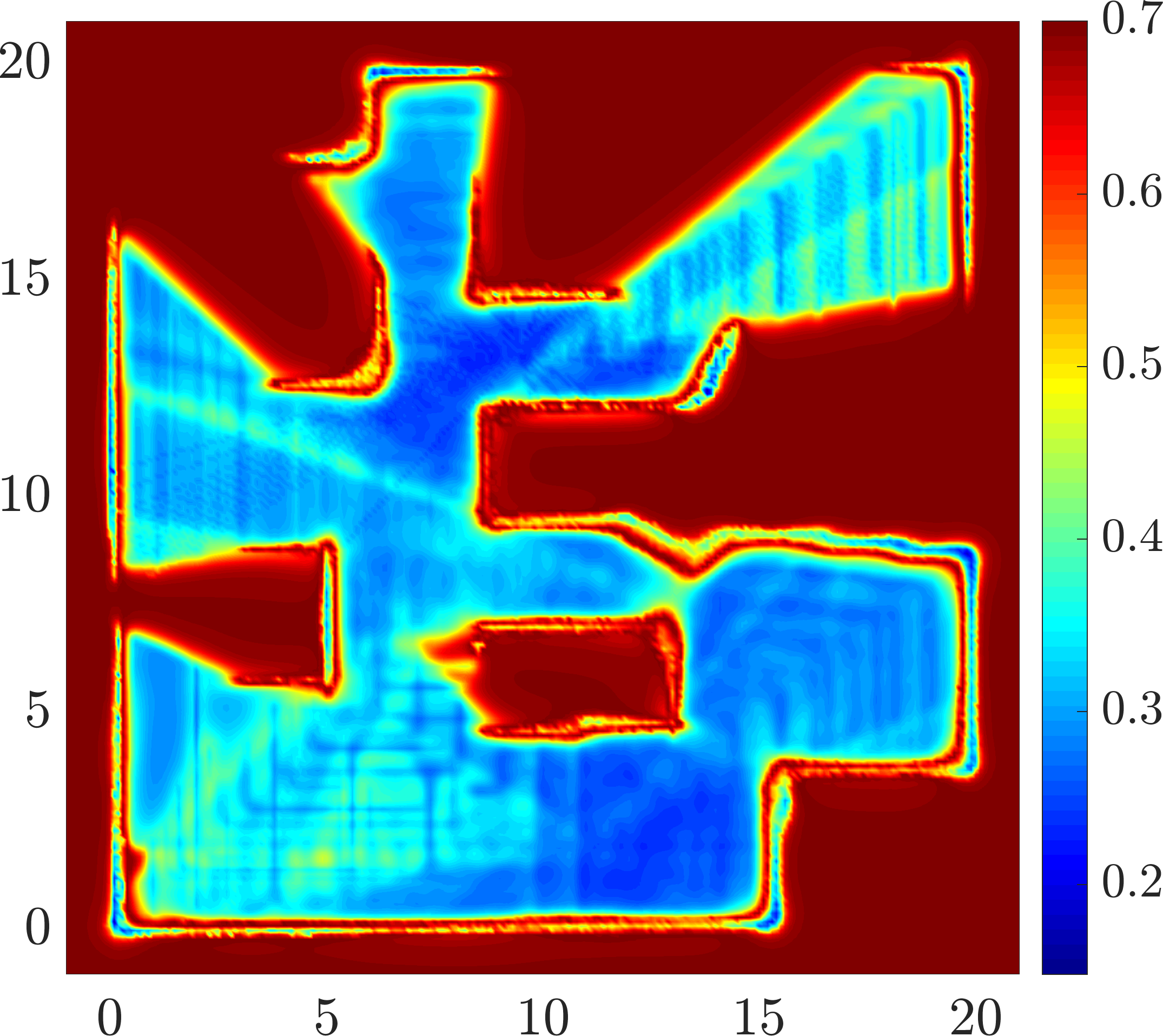}
  \label{fig:ex_MIsurf}}
  \caption{Inferred continuous occupancy map (left); associated probabilistic frontier map (middle); and mutual information surface (right, discussed in Section~\ref{sec:miexp}). The frontier map highlights the informative regions for further exploration by assigning higher probabilities to frontier points. The lower probabilities show the obstacles and walls while the values greater than the \emph{no discrimination} probability, $0.5$, can be considered as frontiers. In the MI surface, the areas beyond the current perception field of the robot preserve their initial entropy values and the higher values demonstrate regions with greater information gain. The map dimensions are in meters and the MI values in nats.}
  \label{fig:ex_maps}
\end{figure}

The frontier surface is converted to a probability frontier map through the incorporation of the map uncertainty. To squash the frontier and variance values into the range $[0, 1]$, a logistic regression classifier with inputs from $\bar{f}^{[i]}$ and map uncertainty $\sigma^{[i]}$ is applied to data which yields
\begin{equation}
\label{eq:logisticf}
	\vspace{-0.1cm}
	p(f^{[i]}|m^{[i]}, w_f^{[i]}) = \frac{1}{1+\exp(-w_f^{[i]} \bar{f}^{[i]})}
\end{equation}
where $w_f^{[i]} = \gamma_f \sqrt{\lambda^{[i]}}$ denotes the required weights, $\lambda^{[i]} \triangleq \sigma_{min} / {\sigma^{[i]}}$ is the bounded information associated with location $i$, and $\gamma_f > 0$ is a constant to control the sigmoid shape. The details of the frontier map computations are presented in Algorithm~\ref{alg:frontier}. Figure~\ref{fig:ex_maps} (middle) depicts an instance of the frontier map from an exploration experiment in the Cave environment~\citep{Radish_data_set}.

In practice, the following steps are required to use the frontier map and check the termination condition:
  \begin{enumerate}
\item The probabilistic frontier map is converted to a binary map using a pre-defined threshold. Note that any point with a probability higher than $0.5$ is potentially a valid frontier.
\item The binary map of frontiers is clustered into subsets of candidate macro-actions.
\item The centroids of clusters construct a discrete action set at time-step $t$, i.e.\@ $\mathcal{A}_t$, that is used in the utility maximization step.
\item The robot plans a path to each centroid (macro-action) to check its reachability. A centroid that is not reachable is then removed from the action set.
\item The exploration mission continues until the action set $\mathcal{A}_t$ is not empty (repeats from step 1).
\end{enumerate}

\begin{algorithm}[t]
\caption[BuildFrontierMap]{\texttt{BuildFrontierMap}()}
\label{alg:frontier}
\begin{algorithmic}[1]
\Require Current map $m$, $\boldsymbol\sigma$ and occupied map $m_o$, $\boldsymbol\sigma_o$;
\State // Compute boundaries
\State $dm \gets$  $\lVert\nabla p(m)\rVert_1$, $dm_o \gets$  $\lVert\nabla p(m_o)\rVert_1$
\State $\sigma_{min} \gets \min(\boldsymbol\sigma)$
\State $f \gets \varnothing$
\State // Compute probabilistic frontiers 
\For {all $i \in \mathcal{M}$}
\State $\bar{f}^{[i]} \gets {dm}^{[i]} - \beta({dm}_o^{[i]} + m_o^{[i]} - 0.5)$
\State $w_f^{[i]} \gets \gamma_f\ \mathrm{sqrt}(\sigma_{min} / \boldsymbol\sigma^{[i]})$ // Logistic regression weights
\State $f^{[i]} \gets (1 + \exp(-w_f^{[i]} \bar{f}^{[i]}))^{-1}$ // Squash data into (0,1), Equation~\eqref{eq:logisticf}
\EndFor
\Return $f$
\end{algorithmic}
\end{algorithm}

\subsection{Computational complexity}
\label{subsec:timecomplex}

For the mapping algorithms, the computational cost of GPs is $\bigO{n^3_t}$, given the need to invert a matrix of the size of training data, $n_t = n_o + n_f$. BCM scales linearly with the number of map points, $n_m$. The overall map update operation involves a nearest neighbor query for each test point, $n_q$, and the logistic regression classifier is at worst linear in the number of map points resulting in $\bigO{n_{t}^3 + n_{q} \log n_{q} + n_m}$.

A more sophisticated approximation approach can reduce the computational complexity further. The fully independent training conditional (FITC)~\citep{snelson2006sparse} based on inducing conditionals suggests an $\bigO{n_t n_i^2}$ upper bound where $n_i$ is the number of inducing points. More recently, in~\cite{hensman2013gaussian}, the GP computation upper bound is reduced to $\bigO{n_i^3}$ which brings more flexibility in increasing the number of inducing points.

\section{Exploration}
\label{sec:Exploration}
In the context of autonomous robotic mapping, typically, the main goal is map completion while maintaining the localization accuracy at an reasonable level\footnote{The required localization accuracy is subject to the specific application.}.  
Let $a_t$ be an action from the set of all possible actions $\mathcal{A}_t$ at time $t$. The goal is to choose the action that optimizes the desired objective function. In the following, we define the most common utility functions for the single robot exploration case.

\textbf{Nearest Frontier.} The nearest frontier policy drives the robot towards the closest frontier to its current pose. Geometric frontiers can be extracted from the occupancy map~\citep{keidar2013efficient}. For the GPOM technique, we use the probabilistic frontier map. Let $\mathcal{F}_t$ be the finite set of all detected frontiers at time $t$. Let the action $a_t$ be the planned path from the current robot pose to the frontier $f_t$. The cost function, $f_c:\mathcal{A}_t\rightarrow \mathbb R_{\geq0}$, is the length of the path from the current robot pose to the corresponding frontier. Therefore,
\begin{equation}	
\label{eq:nfcost}	
	a^\star_t = \underset{a_t \in \mathcal{A}_t}{\operatorname{argmin}} \ f_c(a_t)
\end{equation}
In practice, frontier cells/points are clustered, and only those with the size above a threshold are valid. The centroid of each cluster is considered as the target point for path planning.
\begin{remark}
 In general, the path length can be seen as the line integral of the curve with the current robot pose and the frontier as its end points. Thus, one can define a scalar field over the map and calculate the cost as the line integral of the scalar field using a Riemann sum. In~\eqref{eq:nfcost} the integrand is simply $1$.
\end{remark}


\textbf{Information Gain.} Let $f_I(a_t)$, $f_I:\mathcal{A}_t\rightarrow \mathbb R_{\geq0}$, be a function that quantifies the information quality of action $a_t$. To find the action that maximizes the information gain-based utility function, the problem can be written as
\begin{equation}
\label{eq:igutil}
  a^\star_t = \underset{a_t \in \mathcal{A}_t}{\operatorname{argmax}} \ f_I(a_t)
\end{equation}
In other words, the robot takes the action that leads to the maximum return of information. However, as it is evident from~\eqref{eq:igutil} the cost of taking that action is not included in the utility function.


\textbf{Cost-Utility Trade-off.} The third approach is based on the idea of a trade-off between the cost and utility of an action, i.e.\@ the payoff. The total utility function can be constructed by combination of~\eqref{eq:nfcost} and~\eqref{eq:igutil}. The primary problem is that the units of utility/cost functions are different. One solution is to express the cost in the form of information loss (uncertainty). Another approach is to combine them using appropriate coefficients, e.g.\@ a linear combination of the utility and cost functions. Let $g:{\mathbb R}^2_{\geq0}\rightarrow \mathbb R_{\geq0}$ be a function that takes $f_c(a_t)$ and $f_I(a_t)$ as its input arguments. The problem of maximizing the total utility function, \mbox{$u(a_t) \triangleq g(f_I(a_t), f_c(a_t))$}, can then be defined as follows.
\begin{equation}
\label{eq:totutil}
  a^\star_t = \underset{a_t \in \mathcal{A}_t}{\operatorname{argmax}} \ u(a_t)
\end{equation}

\subsection{Mutual information algorithm}
\label{sec:miexp}

MI is the reduction in uncertainty of a random variable due to the knowledge of another random variable~\cite{cover2012elements}. In other words, given a measurement $Z=\boldsymbol z$ from $\mathcal{Z}$ what will be the reduction in the map $M= m$ uncertainty?
The MI between the map and the future measurement $Z_{t+1}=\hat{\boldsymbol z}$ is
\begin{align}
	\nonumber I(M;Z_{t+1}|\boldsymbol z_{1:t}) &= \nonumber \int_{\hat{\boldsymbol z}\in \mathcal{Z}} \sum_{m\in\mathcal{M}} p(m,\hat{\boldsymbol z}|\boldsymbol z_{1:t})\log{\dfrac{p(m,\hat{\boldsymbol z}|\boldsymbol z_{1:t})}{p(m|\boldsymbol z_{1:t}) p(\hat{\boldsymbol z}|\boldsymbol z_{1:t})}} d\hat{\boldsymbol z}
	\\ &= H(M|\boldsymbol z_{1:t}) - \overline{H}(M|Z_{t+1},\boldsymbol z_{1:t})
\label{MI}
\end{align}
where $H(M|\boldsymbol z_{1:t})$ and $\overline{H}(M|Z_{t+1},\boldsymbol z_{1:t})$ are map and map conditional entropy respectively, which by definition are
\begin{equation}	
\label{mapEnt}	
	H(M|\boldsymbol z_{1:t}) = -\sum_{m\in\mathcal{M}} p(m|\boldsymbol z_{1:t}) \log{p(m|\boldsymbol z_{1:t})}
\end{equation}
\begin{equation}
	\overline{H}(M|Z_{t+1},\boldsymbol z_{1:t}) = \int_{\hat{\boldsymbol z}\in \mathcal{Z}} p(\hat{\boldsymbol z}|\boldsymbol z_{1:t}) H(M|Z_{t+1}=\hat{\boldsymbol z},\boldsymbol z_{1:t}) d\hat{\boldsymbol z}
\label{eq:condEnt}
\end{equation}
To compute the map conditional entropy, the predicted map posterior given the new measurement \mbox{$Z_{t+1}=\hat{\boldsymbol z}_{t+1}$} is required. The Bayesian inference finds the posterior probability for each map point $m^{[i]}$ and $k$-th beam of the range-finder as
\begin{equation}	
\label{mapPosterior}	
	p(m^{[i]}|\hat{\boldsymbol z}^{[k]}_{t+1}, \boldsymbol z_{1:t}) = \frac{p(\hat{\boldsymbol z}^{[k]}_{t+1}|m^{[i]}) p(m^{[i]}|\boldsymbol z_{1:t})}{p(\hat{\boldsymbol z}^{[k]}_{t+1}|\boldsymbol z_{1:t})}
\end{equation}
\begin{equation}	
\label{marginalLikelihood}	
	p(\hat{\boldsymbol z}^{[k]}_{t+1}|\boldsymbol z_{1:t})=\sum_{m^{[i]}\in\mathcal{M}} p(\hat{\boldsymbol z}^{[k]}_{t+1}|m^{[i]}) p(m^{[i]}|\boldsymbol z_{1:t})
\end{equation}
The likelihood function \mbox{$p(\hat{\boldsymbol z}^{[k]}_{t+1}|M=m^{[i]})$} is a beam-based mixture measurement model, where the term \mbox{$p(\hat{\boldsymbol z}^{[k]}_{t+1}|M=0)$} can be interpreted as the likelihood of not observing the map point at location $i$, i.e.\@ uniform distribution. The term $p(\hat{\boldsymbol z}^{[k]}_{t+1}|\boldsymbol z_{1:t})$ is the marginal distribution over measurements which is denoted by $p_z$ in line~\ref{line:mipz} of Algorithm~\ref{alg:mi}. By numerically integrating over a desired beam range, we can compute the predicted map posterior entropy using Equation~\eqref{eq:condEnt}. Note that the conditional entropy does not depend on the realization of future measurements, but it is an average over them.
\begin{algorithm}[t!]
\caption{\texttt{BuildMIMap}()}
\label{alg:mi}
\begin{algorithmic}[1]
\Require Robot pose or desired location, current map estimate $m$, numerical integration resolution $s_z$, sensor model;
\State $\bar{m} \gets m$
\State // Initialize MI map using current map entropy 
\State $I \gets -(m \log(m) + (1- m) \log(1- m))$
\For {all $k$} // Loop over all sensor beams
\State Compute $\hat{z}^{[k]}_{t+1}$ and $\mathcal{I}^{[k]}_{t+1}$ using ray casting in $m$
\State // Calculate map conditional entropy along beam $k$
\For {$i \in \mathcal{I}^{[k]}_{t+1}$}
\State $\bar{h} \gets 0$ // Initialize map conditional entropy
\State $z \gets s_z^{-1}$ // Initialize range dummy variable
\While {$z \leq \hat{z}^{[k]}_{t+1}$}
\State // Calculate marginal measurement probability $p_z$, Equation~\eqref{marginalLikelihood}
\State $p_1 \gets p(z|M=0)$ 
\State $p_2 \gets 0$
\For {$j \in \mathcal{I}^{[k]}_{t+1}$}
\State $p_1 \gets p_1 (1-m^{[j]})$
\State \begin{varwidth}[t]{\columnwidth} 
        $p_2 \gets p_2 +$ \par 
        \hskip\algorithmicindent $p(z|M=m^{[j]}) m^{[j]} \displaystyle\prod_{l < j}{(1-m^{[l]})}$
       \end{varwidth}
\EndFor
\State $p_z \gets p_1 + p_2$  \label{line:mipz}
\State // Map prediction at point $i$ along beam $k$
\State \begin{varwidth}[t]{\columnwidth}  
	$\bar{m}^{[i]} \gets p_z^{-1} p(z|M=m^{[i]})$
        $m^{[i]} \displaystyle\prod_{l < i}{(1-m^{[l]})}$
       \end{varwidth}
\State \begin{varwidth}[t]{\columnwidth} 
        $\bar{h} \gets \bar{h} +$ \par 
        \hskip\algorithmicindent $p_z [\bar{m}^{[i]} \log(\bar{m}^{[i]}) + (1-\bar{m}^{[i]}) \log(1-\bar{m}^{[i]})]$
       \end{varwidth}
\State $z \gets z + s_z^{-1}$ // Increase range along the beam
\EndWhile
\State $I^{[i]} \gets I^{[i]} + \bar{h} s_z^{-1}$ // Equation~\eqref{eq:updatemi}
\EndFor
\EndFor
\Return $I$
\end{algorithmic}
\end{algorithm}

Let $\mathcal{I}^{[k]}_{t+1}$ be the index set of map points that are in the perception field of the $k$-th sensor beam at time $t+1$. At any robot location, $\forall i \in \mathcal{I}^{[k]}_{t+1}$, the MI can be written as
\begin{equation}
\label{eq:updatemi}
	I^{[i]} = h(m^{[i]}) - \overline{h}(m^{[i]})
\end{equation}
where $h(m^{[i]})$ is the current entropy of the map point $m^{[i]}$ and $\overline{h}(m^{[i]})$ is the estimated map conditional entropy. In practice, at each time-step, the map is initialized with the current map entropy, $H(M|\boldsymbol z_{1:t})$, and for all map points inside the current perception field the estimated map conditional entropy is subtracted from corresponding initial values. In Algorithm~\ref{alg:mi}, the implementation of the MI map is given where $s_z$ denotes the numerical resolution of integration. In Figure~\ref{fig:ex_maps}, an estimated MI map during an exploration experiments in the Cave environment~\citep{Radish_data_set} is depicted.

\textbf{Computational Complexity.} For MI surface, the time complexity is at worst quadratic in the number of map points in the current perception field of the robot, $n_p = \lvert \bigcup_{k=1}^{n_z} \mathcal{I}^{[k]}_{t+1} \rvert$, and linear in the number of sensor beams, $n_z$, and numerical integration's resolution, $s_z$, resulting in $\bigO{n^2_p n_z s_z}$.

\subsection{Decision making}
\label{sec:midc}
Let each geometric frontier be regarded as a macro-action. The action space can thus be defined as \mbox{$\mathcal{A}_t = \{a_t^{[j]}\}^{n_a}_{j=1}$}. We define the utility function as the difference between the total expected information gain predicted at the macro-action $a_t$, $f_I(a_t)$, and the corresponding path length from the current robot pose to the same macro-action, $f_c(a_t)$, as follows
\begin{equation}
\label{totMI}	
	f_I(a_t) \triangleq \sum_{k=1}^{n_z} \sum_{i \in \mathcal{I}^{[k]}} I^{[i]}(a_t)
\end{equation}
\begin{equation}
\label{utility}	
	u(a_t) \triangleq \alpha f_I(a_t) - f_c(a_t)
\end{equation}
where $\alpha$ is a factor to relate information gain to the cost of motion. Note that the expectation over future measurements and path lengths is already incorporated into the information and cost functions. 

The optimal action $a_t^\star$ directs the robot towards the frontier with the best balance between information gain and travel cost. This greedy action selection is similar to what is known as next-best view planning in the literature~\cite{gonzalez2002navigation,surmann2003autonomous}.
 
\subsection{Map regeneration}
\label{subsec:regeneration}

Loop closure during SLAM can change the map significantly. To account for such changes, we reset and learn the occupancy map with all the available data again. To be able to efficiently detect such a drift in the GPOM we measure the Jensen-Shannon Divergence (JSD)~\citep{lin1991divergence}. The generalized JSD for $n$ probability, $p_1,p_2,...,p_n$, with weights $\pi_1,\pi_2,...,\pi_n$ is
\begin{equation}	
\label{eq:jsdiv}	
	JS_{\pi}(p_1,p_2,...,p_n) = H(\sum_{i=1}^n\pi_ip_i)-\sum_{i=1}^n\pi_iH(p_i)
\end{equation}
where $H(\cdot)$ is the Shannon entropy function and $p(x_i)$ is the probability associated with variable $x_i$. All weights are set uniformly as all points are equal.

Alternatively, cumulative relative entropy by summing the computed Jensen-Shannon entropy in each iteration shows map drifts over a period and contains the history of map variations. Consequently, the method is less sensitive to small sudden changes.
\begin{remark}
 The main advantage of JSD over Kullback-Leibler divergence, in this case, is that JSD is bounded. As a result, it is more suitable for decision making~\citep{lin1991divergence}.
\end{remark}

\section{Results and Discussion}
\label{sec:Results}
We now present results using two publicly available datasets~\citep{Radish_data_set}. In the first scenario, we use the Intel research lab. map which is a highly structured indoor environment. The second scenario is based on the University of Freiburg campus area. The second map is almost ten times larger than the Intel map and is an example of a large-scale environment with open areas.

The experiments include comparison among the original nearest frontier (NF)~\citep{yamauchi1997frontier}, MI-based exploration using OGM (OGMI), the natural extension of NF with a GPOM representation (GPNF) \citep{maani2014com}, and the proposed MI-based (GPMI) exploration approaches. NF and OGMI results are computed using OGMs while for the GPOM-based methods the I-GPOM2 representation and the probabilistic frontier map proposed in this work are employed. For all the techniques, we use the $A^*$ algorithm to find the shortest path from the robot position to any frontier. The path cost is calculated using the Euclidean distance between map points. Details about the compared methods are described in Table~\ref{tab:expmethods}.

\begin{figure}[t]
  \centering 
  \includegraphics[width=.4\columnwidth,trim={1.5cm 1.5cm 1.5cm 1.5cm},clip]{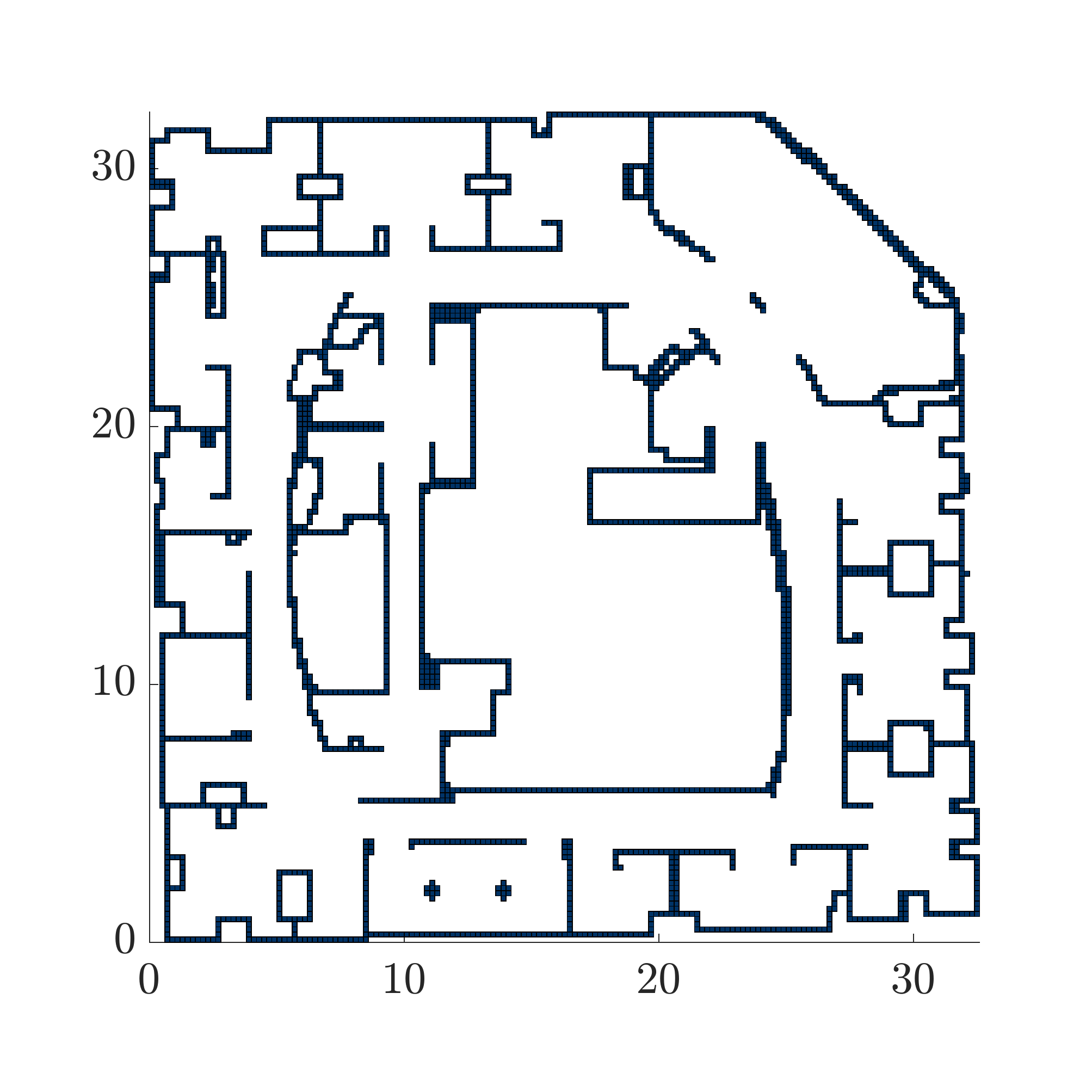}
  \caption{The constructed environment for exploration experiments using the binary map of obstacles from the Intel dataset.}
  \label{fig:intel_obsmap}
  
\end{figure}

\begin{table}[t]
\footnotesize
\centering
\caption{The compared exploration methods and their corresponding attributes.}
\begin{tabular}{lcccc}
\toprule
		& NF		& OGMI 		& GPNF	 	& GPMI 	\\ \midrule

SLAM		& Pose SLAM	& Pose SLAM	& Pose SLAM	& Pose SLAM	\\
Mapping		& OGM		& OGM		& I-GPOM2	& I-GPOM2	\\
Frontiers 	& binary	& binary	& probabilistic	& probabilistic	\\
Utility 	& path length	& MI+path length & path length	& MI+path length \\
Planner		& $A^*$		& $A^*$		& $A^*$		& $A^*$		\\ \bottomrule
\end{tabular}
\label{tab:expmethods}
\end{table}

\begin{table}[t]
\scriptsize
\centering
\caption{Parameters for frontier and MI maps computations. Note that the employed maximum sensor range and the maximum range used in the MI algorithm for prediction do not need to be the same.}
\begin{tabular}{lll}
\toprule
Parameter			& Symbol				 & Value \\ \midrule
\multicolumn{3}{l}{$1)$ Beam-based mixture measurement model:} \\

Hit std				& $\sigma_{hit}$	& 0.03 $\m$	\\
Short decay			& $\lambda_{short}$	& 0.2 $\m$	 	\\
Max range and size of \texttt{TestDataWindow}		& $r_{max}$		& 	\\
$-$ Intel map			& 			& 14.0 $\m$	\\
$-$ Freiburg map		& 			& 60.0 $\m$	\\
Hit weight			& $z_{hit}$		& 0.7		\\
Short weight			& $z_{short}$		& 0.1		\\ 
Max weight			& $z_{max}$		& 0.1	 	\\
Random weight			& $z_{rand}$		& 0.1		\\ \midrule
\multicolumn{3}{l}{$2)$ Frontier map:} \\
Occupied boundaries factor	& $\beta$		& 3.0		\\
Logistic regression weight	& $\gamma$		& 10.0		\\ 
Frontier probability threshold & $-$			& 		\\ 
$-$ Intel map			& 			& 0.6	\\
$-$ Freiburg map		& 			& 0.55	\\ 
Frontier cluster size 		& $-$			& 		\\ 
$-$ Intel map			& 			& 14	\\
$-$ Freiburg map		& 			& 3	\\ 
Number of clusters		& $-$			& 		\\ 
$-$ Intel map			& 			& 20	\\
$-$ Freiburg map		& 			& 5		\\ \midrule
\multicolumn{3}{l}{$3)$ MI map and utility function:} \\
No. of sensor beams over 360 deg 		& $n_z$		& 133		\\
Max range			& $r_{max}$	& 	\\
$-$ Intel map			& 		& 4.0 $\m$	\\
$-$ Freiburg map		& 		& 60.0 $\m$	\\
Numerical integration resolution 		& $s_z$		& 		\\
$-$ Intel map			& 		& 10/3 $\m^{-1}$	\\
$-$ Freiburg map		& 		& 1 $\m^{-1}$	\\
Information gain factor 			& $\alpha$	& 		\\
$-$ Intel map			& 		& 0.1	\\
$-$ Freiburg map		& 		& 0.5	\\
Occupied probability threshold 			& $p_{o}$	& 0.65		\\
Unoccupied probability threshold		& $p_{f}$	& 		\\ 
$-$ Intel map			& 			& 0.35	\\
$-$ Freiburg map		& 			& 0.4	\\	\bottomrule
\end{tabular}
\label{tab:param}
\end{table}

\begin{figure}[t!]
  \centering  
  \subfloat[]{
    \includegraphics[width=0.345\columnwidth]{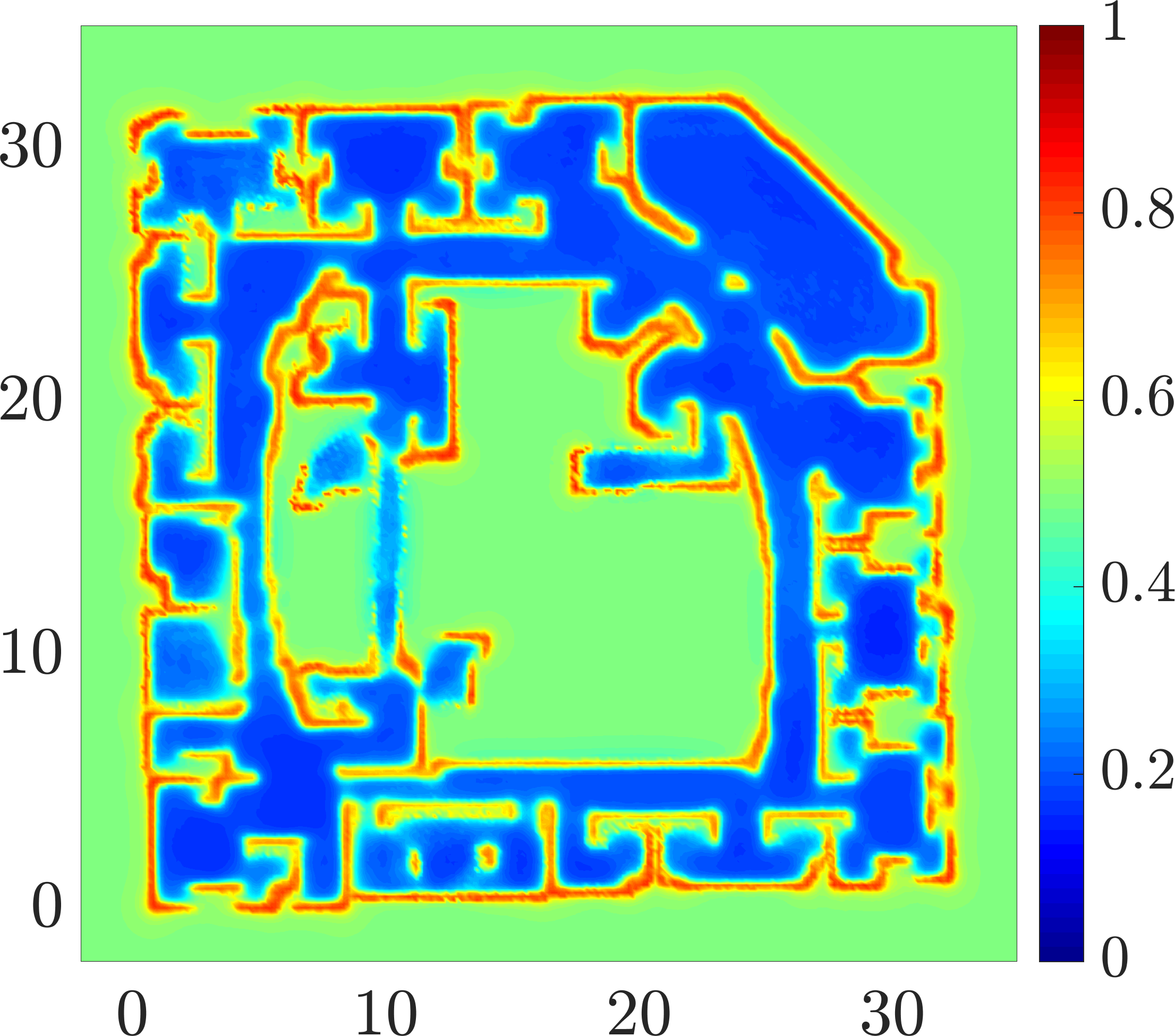}
    \label{fig:intel_gpom}
    }
  \subfloat[]{
    \includegraphics[width=0.3\columnwidth]{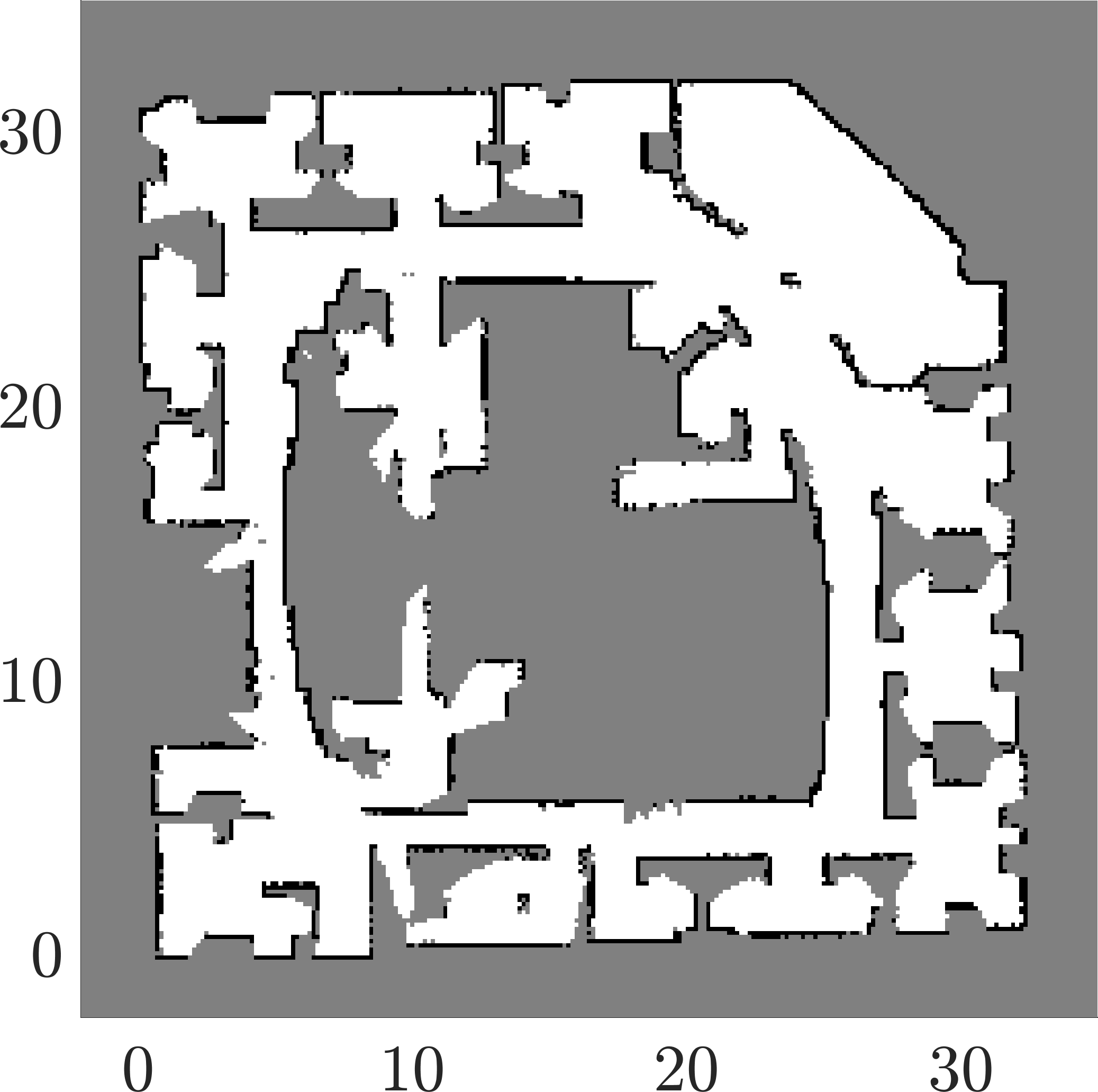}
    \label{fig:intel_ogm}
    }
  \subfloat[]{
    \includegraphics[width=0.345\columnwidth]{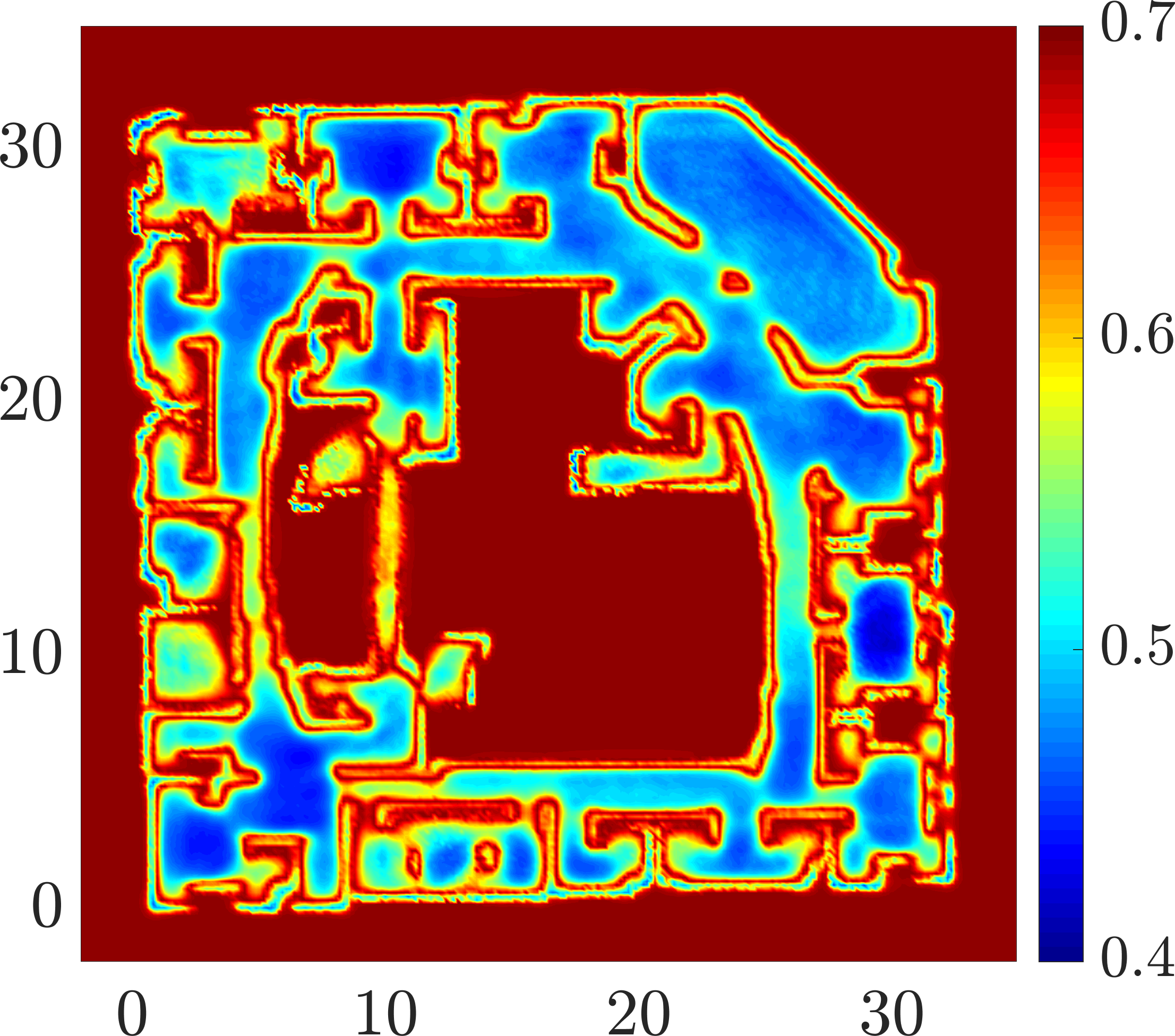}
    \label{fig:intel_ent}
    }
  \caption{MI-based exploration in the Intel map derived from the Intel dataset. (a) I-GPOM2, (b) the equivalent OGM computed at the end of the experiment (c) corresponding entropy map of the GPOM (nats). The sparse observations due to the occluded perception field in a complex environment such as the Intel map signifies the capabilities of OGM and GPOM methods to cope with such limitations. Map dimensions are in meters, and the maps are built with the resolution $0.135\m$.}
  \label{fig:intelResults2}
\end{figure}

\begin{figure}
  \centering
  \includegraphics[width=.4\columnwidth]{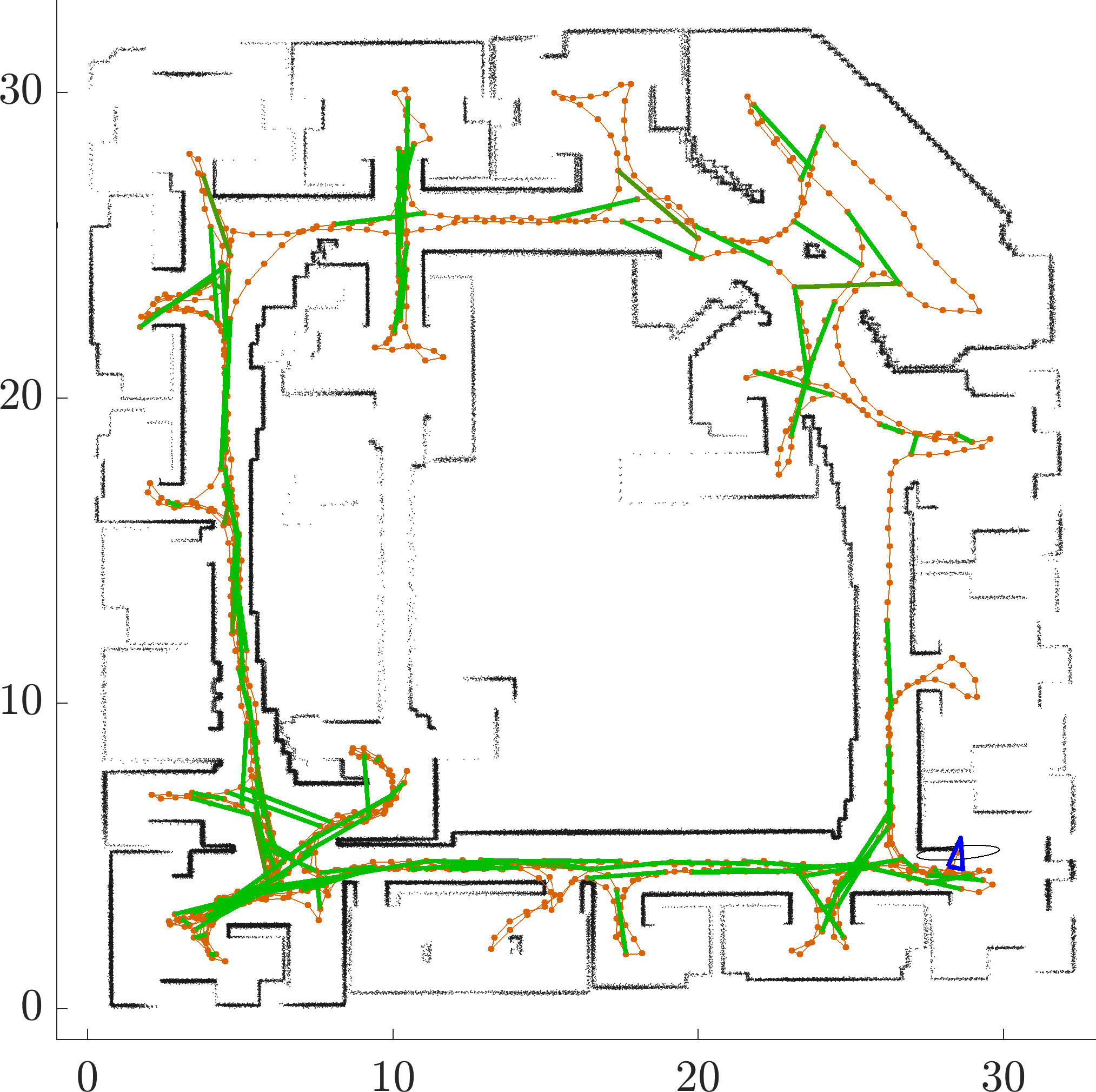}
  \caption{Pose SLAM map of the MI-based exploration in the Intel map derived from the Intel dataset. Dotted (red) curves are the robot path and connecting lines (green) indicate loop-closures. Map dimensions are in meters. The starting robot position is at (18,26), horizontally and vertically, respectively, and the robot terminates the exploration mission at the most bottom right room.}
  \label{fig:intel_slam}
\end{figure}

\begin{figure}[th]
  \centering 
  \includegraphics[width=.6\columnwidth,trim={1.cm 1.cm 1.cm 1.cm},clip]{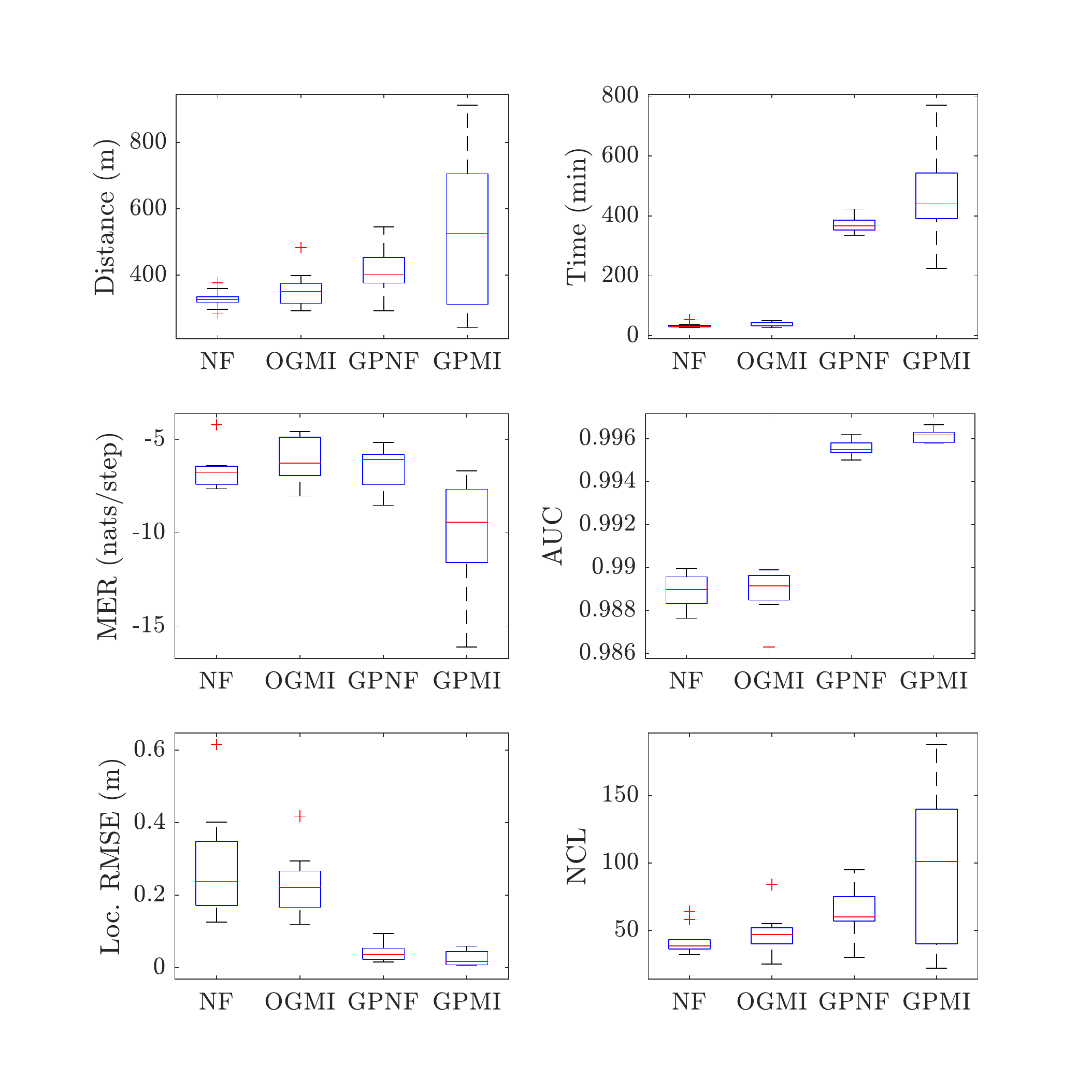}
  \caption{The box plots show comparison of different exploration strategies in the Intel dataset from $10$ independent runs. The compared criteria are travel distance ($\m$), time (min), map entropy rate (nats/step), the mapping performance using the area under the receiving operating characteristic curve, localization root mean-squared error ($\m$), and the number of closed loops by Pose SLAM.}
  \label{fig:intel_boxplot}
\end{figure}

\subsection{Experimental setup}
\label{subsec:setup}
The environment is constructed using a binary map of obstacles and, for the Intel map, is shown in Figure~\ref{fig:intel_obsmap}. The simulated robot is equipped with odometric and laser range-finder sensors to provide the required sensory inputs for Pose SLAM. The odometric and laser range-finder sensors noise covariances are set to \mbox{$\boldsymbol \Sigma_u = \diag(0.1\m, 0.1\m, 0.0026\rad)^2$} and $\boldsymbol \Sigma_y = \diag(0.03\m, 0.03\m, 0.0013\rad)^2$, respectively. The motion of the robot is modeled using a velocity motion model~\citep[Chapter 5]{thrun2005probabilistic} and a proportional control law for following a planned trajectory. Laser beams are simulated through ray-casting operation over the ground truth map using the true robot pose. In all the presented results, Pose SLAM \citep{ila2010information} is included as the backbone to provide localization data together with the number of closed loops. Additionally, for each map, Pose SLAM parameters are set and fixed regardless of the exploration method. 

The localization Root Mean-Squared Error (RMSE) is computed at the end of each experiment by the difference in the robot traveled path (estimated and ground truth poses) to highlight the effect of each exploration approach on the localization accuracy. The required parameters for the beam-based mixture measurement model~\citep{thrun2005probabilistic}, frontier maps, and MI maps computations are listed in Table~\ref{tab:param}. The sensitivity of the parameters in Table~\ref{tab:param} is not high and slight variations of them ($\sim 10\%$) do not affect the presented results.

The implementation has been developed in MATLAB and GP computations have been implemented by modifying the open source GP library in \citet{rasmussen2006gaussian}. As described in Section~\ref{subsec:regeneration}, during exploration, map drifts occur due to loop-closure in the SLAM process. As it is computationally expensive to process all measurements from scratch at each iteration, a mechanism has been adopted to address the problem. The cumulative relative entropy by summing the computed JSD can detect such map drifts.

Each technique is evaluated based on six different criteria, namely, travel distance, mapping and planning time, Map Entropy Rate (MER), AUC of the GP occupancy map calculated at the end of each experiment using all available observations, localization RMSE, and the Number of Closed Loops (NCL). The map entropy at any time-step can be computed using~\eqref{mapEnt}. The map entropy calculation can become independent of the map resolution following the idea in~\citet{stachniss2005information}; that is the cell area, i.e. the squared of the map resolution, weights each entropy term. To see the performance of decision-making across the entire an experiment, the MER is then computed at the end of each experiment using the difference between final and initial map entropies divided by the number of exploration steps. Note that none of the compared exploration strategies explicitly plans for loop-closing actions. For each dataset, the results are from $10$ independent runs using the same setup and parameters.

\begin{figure}[!t]
  \centering 
  \subfloat{\includegraphics[width=.335\columnwidth, trim={0cm 0cm 0cm 0.5cm},clip]{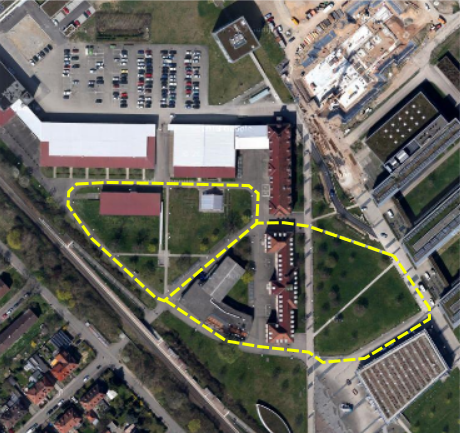}
  \label{fig:frcampus_sat}} 
  \subfloat{\includegraphics[width=.335\columnwidth]{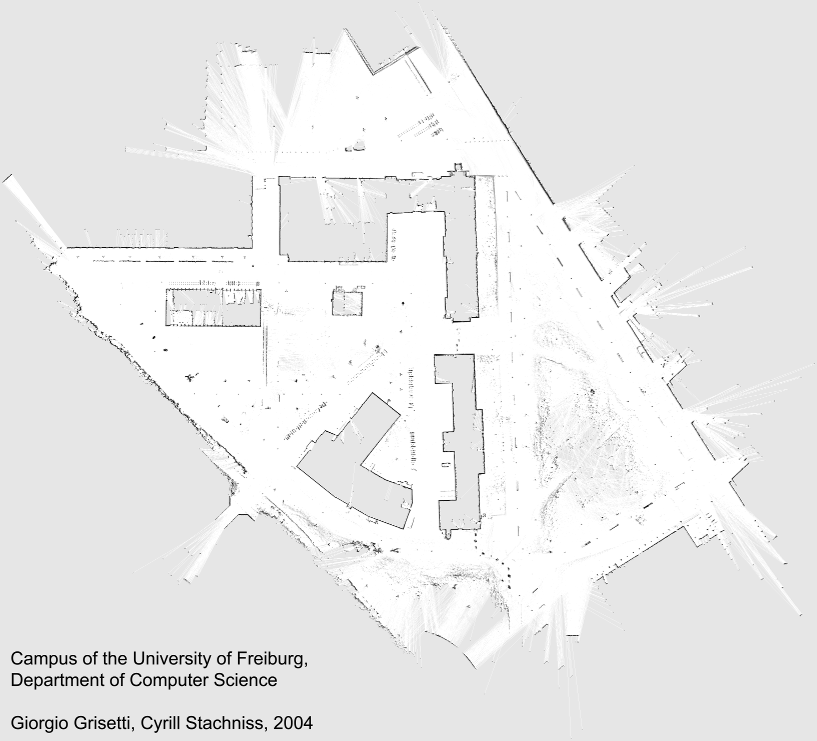}
  \label{fig:frcampus_og}} 
  \subfloat{\includegraphics[width=.32\columnwidth, trim={1.25cm 1.75cm 2.25cm 1.5cm},clip]{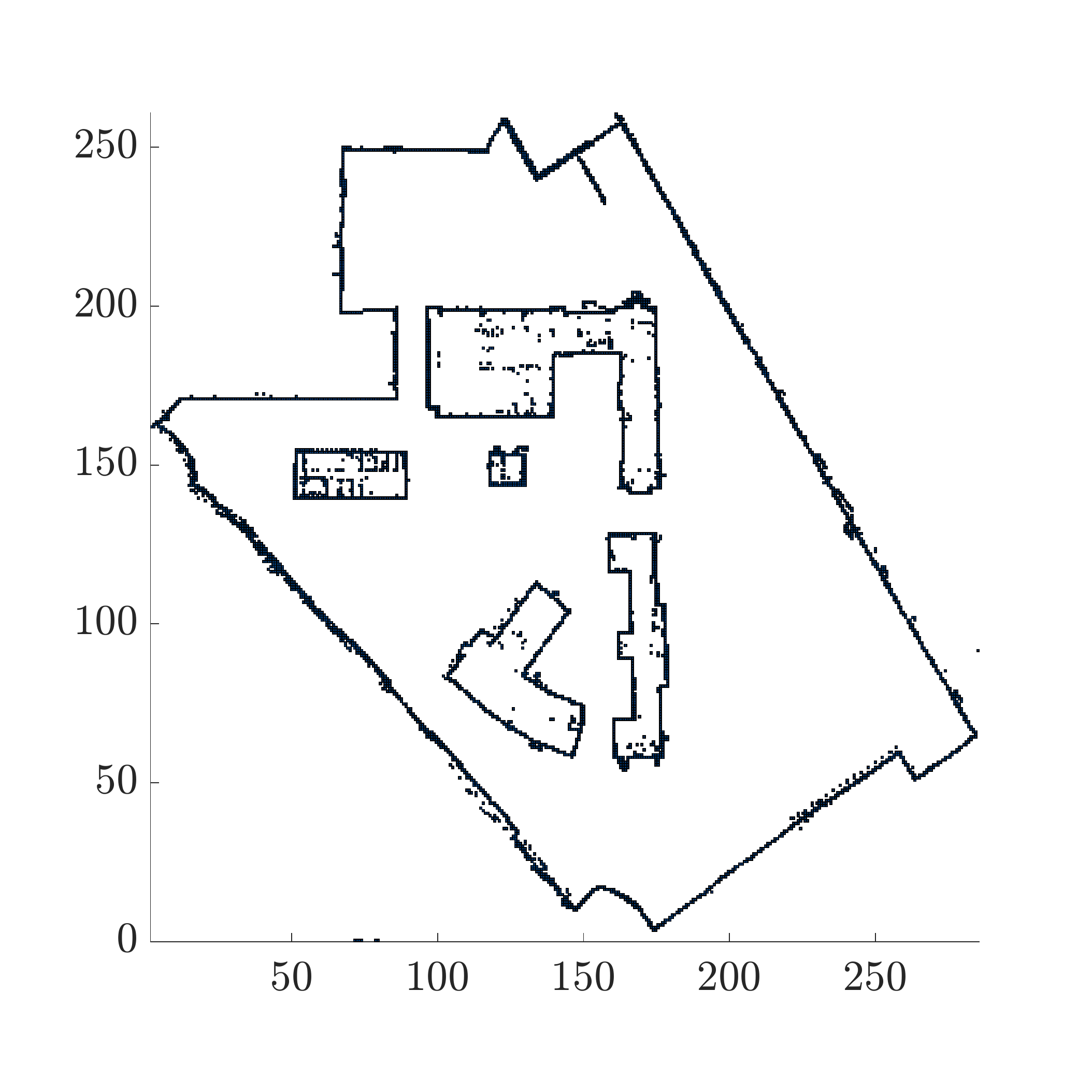}
  \label{fig:frcampus_grid}}
  \caption{The left picture shows the satellite map of the Freiburg University Campus where the yellow dashed line indicate the robot trajectory. The middle figure shows the corresponding occupancy map of the dataset~\citep{Radish_data_set}. The right figure shows the corresponding binary map of obstacles used for exploration experiments. Map dimensions are in meters.}
  \label{fig:frcampus}
  
\end{figure}

\subsection{Exploration results in the Intel map}
An example of the exploration results using GPMI is shown in Figures~\ref{fig:intelResults2} and~\ref{fig:intel_slam}. 
The statistical summary of the results are depicted in Figure~\ref{fig:intel_boxplot}.
The most significant part of the results is related to the map entropy rate in which a negative value means the map entropy has been reduced at each step. In the nearest frontier techniques there is no prediction step regarding map entropy reduction; therefore, the results are purely based on chance and structural shape of the environment. OGMI shows marginal improvements over NF with roughly similar computational times for the exploration mission. Thus, it is the preferred technique in comparison with NF.

GPNF and GPMI exploit I-GPOM2 for mapping, exploration, and planning. GP-based methods handle sparse sensor measurements by learning the structural dependencies (spatial correlations) present in the environment. The significant increase in the map entropy rate is due to this fact. The results from GPMI show higher travel distance and a higher number of closed loops which can be understood from the fact that information gain in the utility function drives the robot to possibly further but more informative targets. As this behavior does not show any undesirable effect on the localization accuracy, it can be concluded that it performs better than the other techniques; however with a higher computational time. The information gain calculation could be sped up by using CSQMI due to its similar behavior to MI~\citep{charrow2015information}. Under the GPMI scheme, the robot chooses macro-actions that balance the cost of traveling and MI between the map and future measurements. Although the utility function does not include the localization uncertainty explicitly, the correlation between robot poses and the map helps to improve the localization accuracy.

\begin{figure}[t]
  \centering 
  \includegraphics[width=.6\columnwidth,trim={1cm 1cm 1cm 1cm},clip]{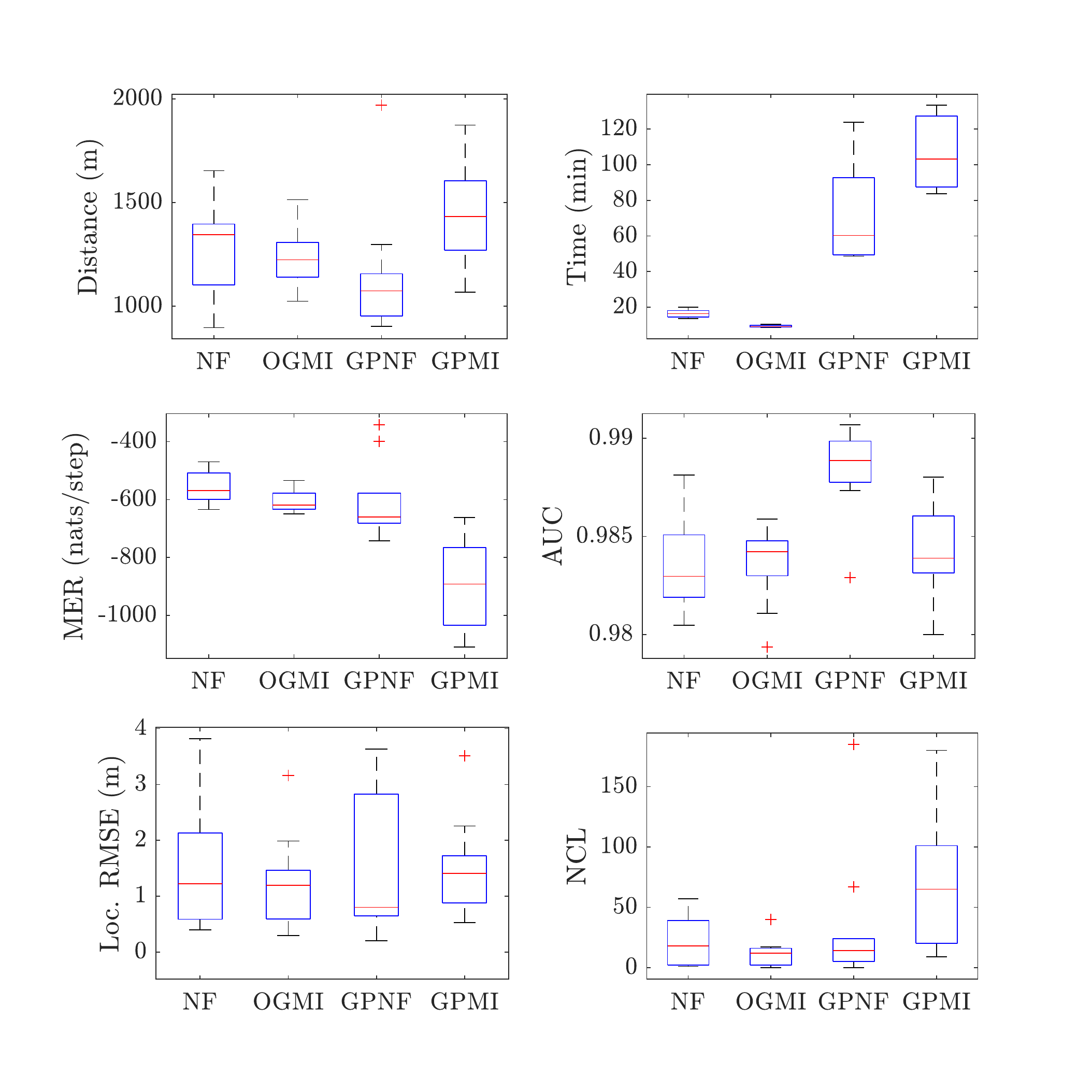}
  \caption{The box plots show comparison of different exploration strategies in the Freiburg campus dataset from $10$ independent runs. The compared criteria are travel distance ($\m$), time (min), map entropy rate (nats/step), the mapping performance using the area under the receiving operating characteristic curve, localization root mean-squared error ($\m$), and the number of closed loops by Pose SLAM.}
  \label{fig:fr_boxplot}
\end{figure}

\subsection{Outdoor scenario: Freiburg Campus}
In the second scenario, the map is an outdoor area with a larger size (almost ten times). Figure~\ref{fig:frcampus} shows the satellite map of the area as well as the trajectory that the robot was driven for data collection. Similar to the first experiment, a binary map of the dataset is constructed and used for exploration experiments. The statistical summary of the results is shown in Figure~\ref{fig:fr_boxplot}. To maintain the computational time manageable, the occupancy maps are built with the coarse resolution of $1 \m$.

Overall, the trend is similar to the previous test, and specifically, the map entropy rate plot shows a significant difference between GPMI and the other techniques. Again, this significant map entropy rate improvement has been achieved without any undesirable effects on the localization accuracy. The sharpness of the localization error distribution can be seen as the reliability and repeatability characteristic of GPMI. Since this map has large open areas relative to the robot's sensing range, it is highly unlikely that the robot closes loops by chance. For the GPMI, the number of closed loops has a higher median which supports the idea of implicit loop-closing actions due to the correlations between the map and the robot pose. However, the NCL distribution has wider tails which does not support its repeatability. The exploration times in this environment is less than those of the previous experiment in the Intel map. We associate the faster map exploration results with the combination of the difference in map resolutions and the open shape of the Freiburg campus map. In contrast, the Intel map is highly structured with narrow hallways and small rooms which require a finer map resolution leading to a higher number of query points. Furthermore, in the Intel map, unlike the Freiburg campus map, a larger maximum range does not help the robot to explore the map faster due to the occlusion problem.

\begin{figure}[t]
  \centering 
  \subfloat{\includegraphics[width=.4\columnwidth]{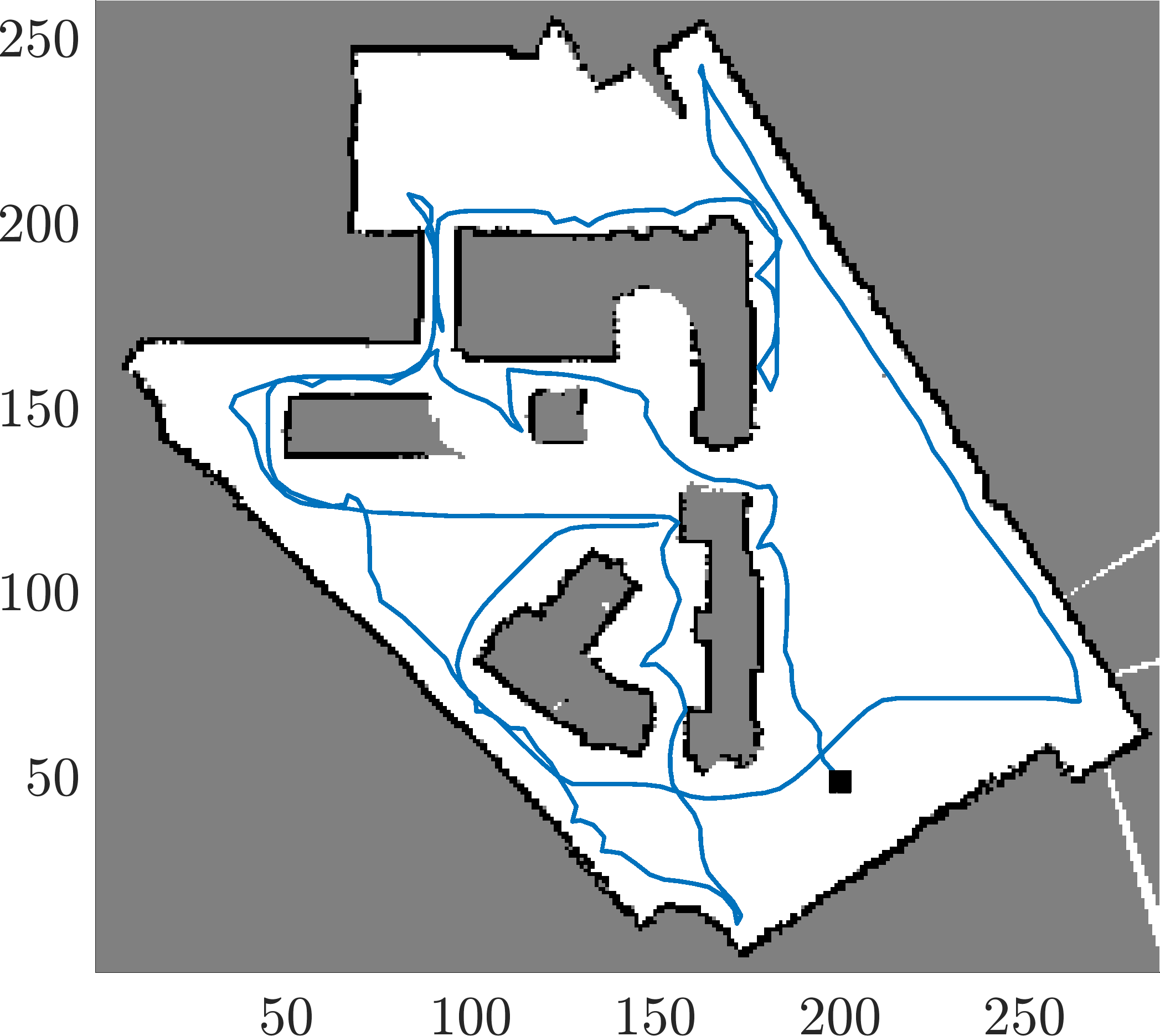}}~
  \subfloat{\includegraphics[width=.4\columnwidth]{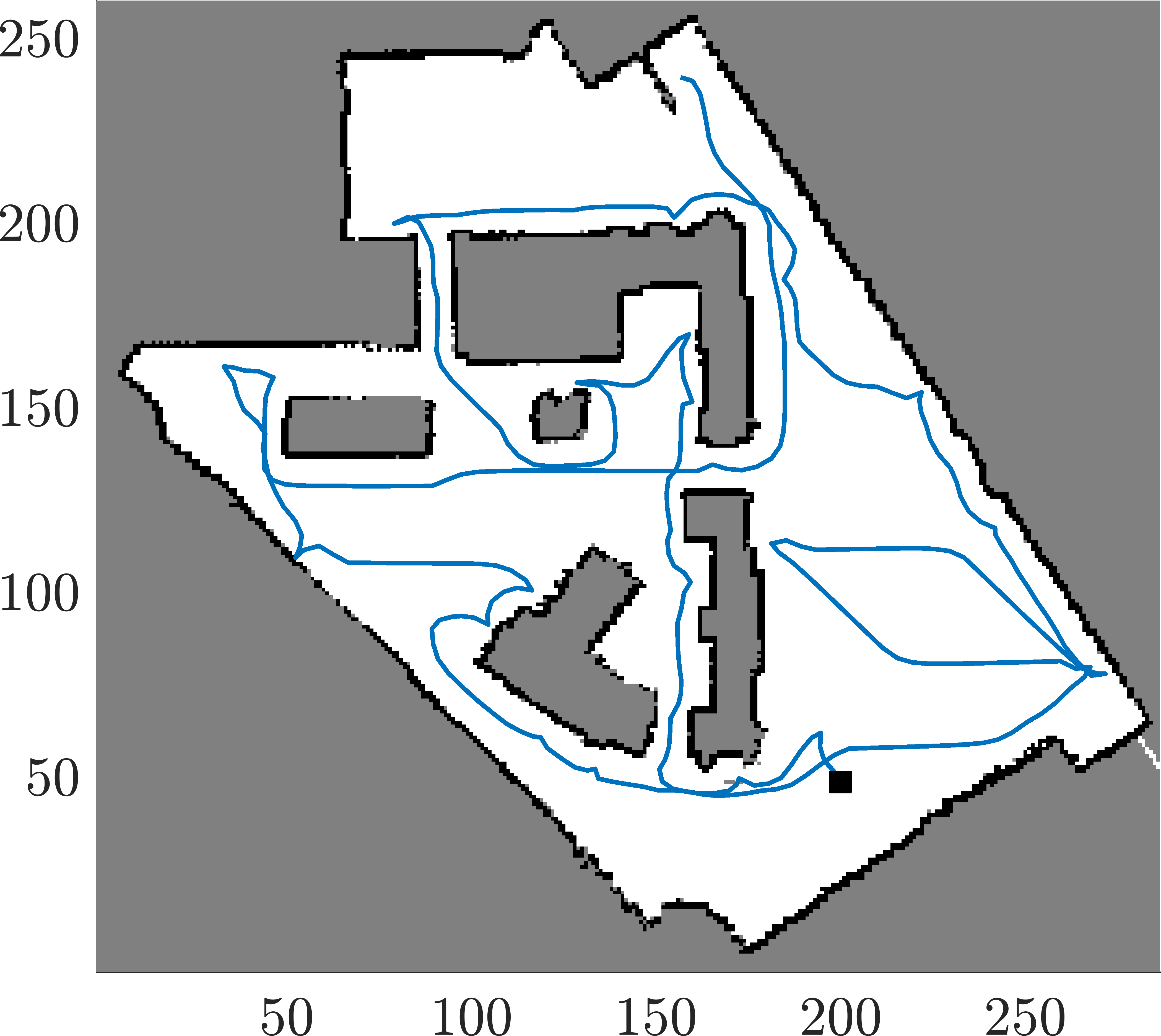}}\\
  \subfloat{\includegraphics[width=.4\columnwidth]{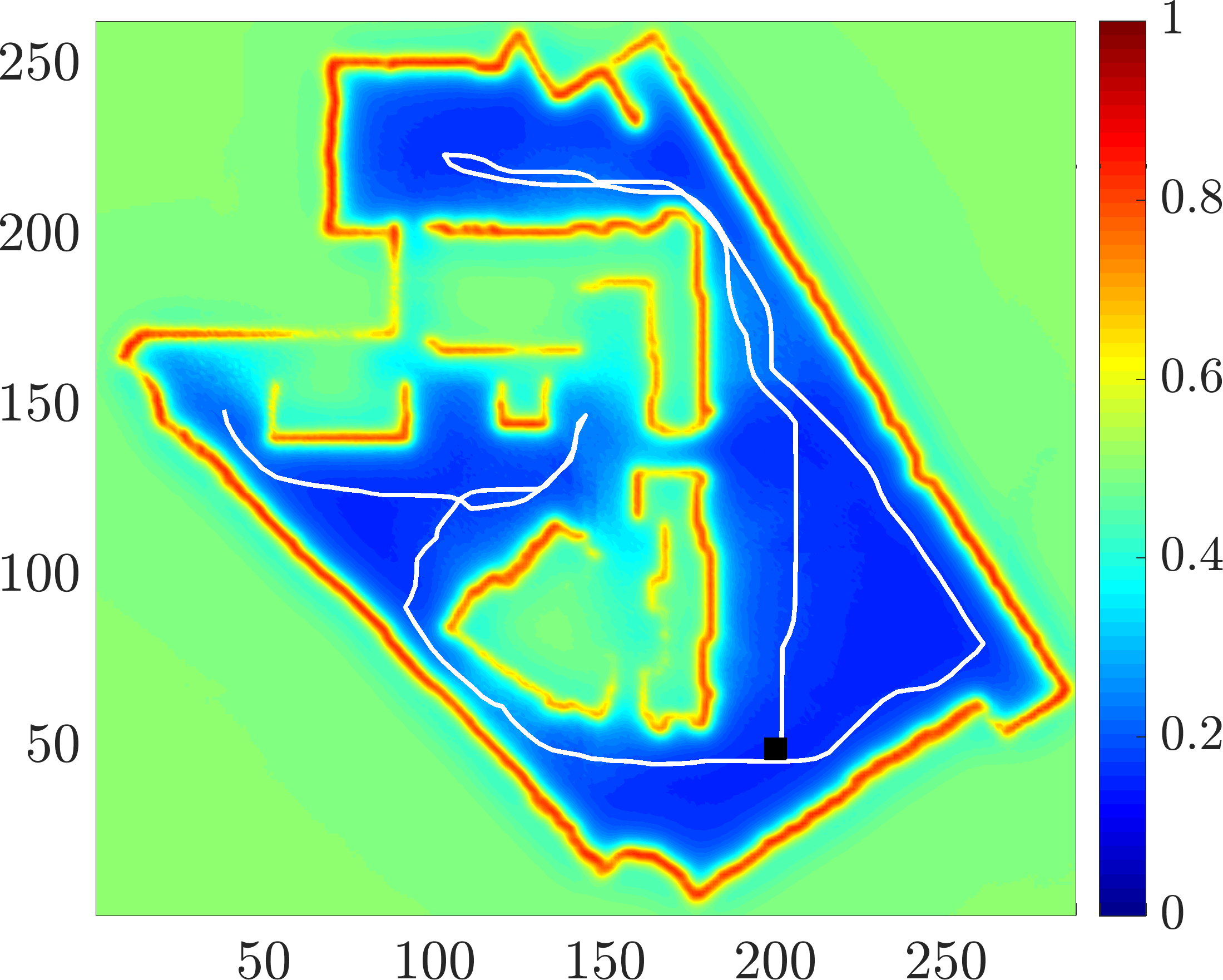}}~
  \subfloat{\includegraphics[width=.4\columnwidth]{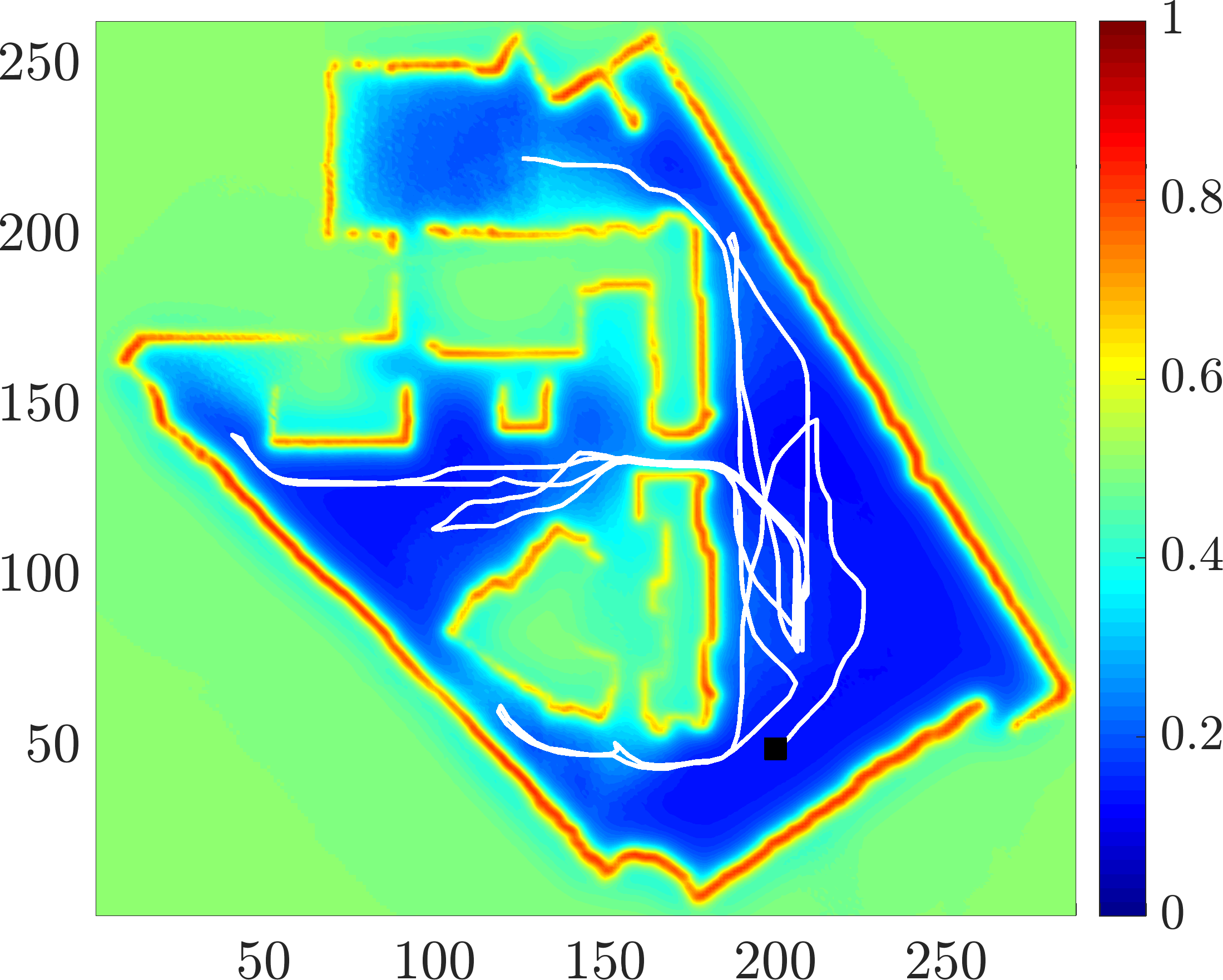}}
  \caption{Illustrative examples of exploration in the Freiburg Campus map. The top left and right, and the bottom left and right figures show the results for NF, OGMI, GPNF, and GPMI, respectively.}
  \label{fig:frcampus_ex}
\end{figure}

Figure~\ref{fig:frcampus_ex} shows the results from an exploration run in Freiburg campus map using NF, OGMI, GPNF, and GPMI. The robot behavior is distinguishable in all four maps. In NF case, the robot tends to travel to every corner in the map to complete the partially observable parts of the map. This behavior leads to trajectories along the boundaries of the map. In OGMI, the prediction of the information gain reduces this effect. However, the OGM requires a higher number of measurements to cover an area; therefore, the robot still needs to travel to the corners. In GPNF case, this effect has been alleviated since the the continuous mapping algorithm can deal with sparse measurements. However, in GPMI case, the robot behaves completely different as by taking the expectation over future measurements (calculating MI) the robot does not act based on the current map uncertainty minimization, but improving the future map state in expectation.

\section{Conclusion and Future Work}
\label{sec:conclusion}
We studied the problem of autonomous mapping and exploration for a range-sensing mobile robot using Gaussian processes maps. The continuity of GPOMs is exploited for a novel representation of geometric frontiers, and we showed that the GP-based mapping and exploration techniques are a competitor for traditional occupancy grid-based techniques. The primary motivations stemmed from the fact that high-dimensional map inference requires fewer observations to infer the map, leading to a faster map entropy reduction. The proposed exploration strategy is based on learning spatial correlations of map points using incremental GP-based regression from sparse range measurements and computing mutual information from the map posterior and conditional entropy. We presented results for two exploration scenarios including a highly structured indoor map as well as a large-scale outdoor area.

When accurate sensors with large coverage relative to the environment are available, existing SLAM techniques can produce reliable localization without the need for an active loop-closure detection. MI-based utility function proposed in this work is suitable for decision making in such scenarios. The more general form of this problem known as active SLAM requires an active search for loop-closures to reduce pose uncertainties. However, the expansion of the state space to both the robot pose and map results in a computationally expensive prediction problem.

Extensions of this work include development of the planning algorithms with longer horizons as well as incorporating the robot pose uncertainty into the mapping and decision-making frameworks~\citep{jadidi2016sampling,jadidi2017warped}. Furthermore, as analyzed and discussed in Subsection~\ref{subsec:timecomplex}, development of computationally more attractive GPOM algorithms remains as an interesting future direction.

\bibliographystyle{SageH}
{\small
\bibliography{references}}

\end{document}

%% file: newcommands.tex
\usepackage{fullpage}

\usepackage[T1]{fontenc}
\usepackage{kpfonts}
\usepackage{libertine}
\usepackage[scaled=0.85]{beramono}

\usepackage{graphicx}
\usepackage{amsmath,amsfonts,amscd, amssymb}
\usepackage{amsthm,thmtools,enumerate,caption}
\usepackage{xspace}
\usepackage{setspace}
\usepackage{perpage}
\usepackage{fancyhdr}
\usepackage{sectsty}
\usepackage{bbm}

\usepackage{algorithm, algorithmicx, algpseudocode}
\usepackage{varwidth}

\let\oldReturn\Return
\renewcommand{\Return}{\State\oldReturn}

\usepackage{booktabs}
\usepackage[caption=false,font=footnotesize]{subfig}

\usepackage[usenames,dvipsnames,svgnames,table]{xcolor}
\definecolor{darkgreen}{rgb}{0.0,0,0.9}

\usepackage[colorlinks=true,pdfpagemode=UseNone,citecolor=OliveGreen,linkcolor=black,urlcolor=BrickRed,
pagebackref]{hyperref}

\newtheorem{theorem}{Theorem}

\theoremstyle{definition}
\newtheorem{definition}{Definition}

\newtheorem{assumption}[theorem]{Assumption}

\newtheorem{remark}{Remark}

\usepackage[numbers]{natbib}
\bibpunct{(}{)}{;}{}{}{,}
\usepackage{multicol}

\newcommand{\diag}{\mathop{\mathrm{diag}}}
\newcommand{\rad}{\mathop{\mathrm{rad}}}
\newcommand{\m}{\mathop{\mathrm{m}}}

\newcommand{\bigO}[1]{\mathcal{O}(#1)} 
\newcommand{\EV}[1]{\mathbb{E}[#1]} 
\newcommand{\Var}[1]{\mathbb{V}[#1]} 
\newcommand{\Cov}[1]{\mathbb{C}\mathbbm{ov}[#1]} 

\algrenewcommand\textproc{}%


\usepackage{enumitem}